\documentclass[11pt]{article}

\usepackage{fullpage}
\usepackage{thmtools}
\usepackage{thm-restate}
\usepackage{graphicx}
\usepackage{float}
\usepackage{booktabs} 
\usepackage{wrapfig}   
\usepackage[table]{xcolor}% for wrapfigure
\usepackage{subcaption}
\usepackage{placeins}
\usepackage{adjustbox}            % To adjust table width
\usepackage{multirow} 
\usepackage{makecell}

\usepackage{color-edits}
\usepackage{enumerate}
\addauthor{ev}{purple}
\addauthor{ishani}{teal}
\addauthor{mika}{green}

\usepackage{amsmath,amssymb,amsfonts,mathtools,amsthm}
\usepackage{bbm}
\usepackage{color}
\usepackage{natbib}

\usepackage[algo2e,linesnumbered,ruled,vlined]{algorithm2e}
\usepackage{algorithmic}
\usepackage{caption}
\usepackage{fancyhdr}
\usepackage{enumitem, comment}
\usepackage{soul} %highlight and underline
 \usepackage{dsfont}
\usepackage[utf8]{inputenc} % allow utf-8 input
\usepackage{csquotes}
\usepackage{fnpct} %for footnotes
\usepackage{float}
\usepackage[utf8]{inputenc}
\usepackage[english]{babel}
\usepackage{xspace}    % Smart spacing with \xspace % to define text macros
\usepackage{nicefrac}

\usepackage{mdframed}
\newmdtheoremenv{theomd}{Theorem}

\newtheorem{theorem}{Theorem}[section]
\newtheorem*{theorem*}{Theorem}

\newtheorem*{claim*}{Claim}

\newtheorem*{proposition*}{Proposition}
\newtheorem{lemma}[theorem]{Lemma}
\newtheorem*{lemma*}{Lemma}
\newtheorem{corollary}[theorem]{Corollary}
\newtheorem{conjecture}[theorem]{Conjecture}
\newtheorem*{conjecture*}{Conjecture}

\newtheorem*{hypothesis*}{Hypothesis}

\theoremstyle{definition}
\newtheorem{definition}[theorem]{Definition}
\newmdtheoremenv{defmd}[theorem]{Definition}

\newtheorem{remark}[theorem]{Remark}

\usepackage{prettyref}
\newcommand{\savehyperref}[2]{\texorpdfstring{\hyperref[#1]{#2}}{#2}}

\newrefformat{eq}{\savehyperref{#1}{\textup{(\ref*{#1})}}}
\newrefformat{def}{\savehyperref{#1}{Definition~\ref*{#1}}}
\newrefformat{thm}{\savehyperref{#1}{Theorem~\ref*{#1}}}
\newrefformat{cor}{\savehyperref{#1}{Corollary~\ref*{#1}}}
\newrefformat{cha}{\savehyperref{#1}{Chapter~\ref*{#1}}}
\newrefformat{sec}{\savehyperref{#1}{Section~\ref*{#1}}}
\newrefformat{app}{\savehyperref{#1}{Appendix~\ref*{#1}}}
\newrefformat{tab}{\savehyperref{#1}{Table~\ref*{#1}}}
\newrefformat{fig}{\savehyperref{#1}{Figure~\ref*{#1}}}
\newrefformat{hyp}{\savehyperref{#1}{Hypothesis~\ref*{#1}}}
\newrefformat{alg}{\savehyperref{#1}{Algorithm~\ref*{#1}}}
\newrefformat{sdp}{\savehyperref{#1}{SDP~\ref*{#1}}}
\newrefformat{rmk}{\savehyperref{#1}{Remark~\ref*{#1}}}
\newrefformat{item}{\savehyperref{#1}{Item~\ref*{#1}}}
\newrefformat{step}{\savehyperref{#1}{step~\ref*{#1}}}
\newrefformat{conj}{\savehyperref{#1}{Conjecture~\ref*{#1}}}
\newrefformat{fact}{\savehyperref{#1}{Fact~\ref*{#1}}}
\newrefformat{prop}{\savehyperref{#1}{Proposition~\ref*{#1}}}
\newrefformat{claim}{\savehyperref{#1}{Claim~\ref*{#1}}}
\newrefformat{relax}{\savehyperref{#1}{Relaxation~\ref*{#1}}}
\newrefformat{red}{\savehyperref{#1}{Reduction~\ref*{#1}}}
\newrefformat{part}{\savehyperref{#1}{Part~\ref*{#1}}}
\newrefformat{prob}{\savehyperref{#1}{Problem~\ref*{#1}}}
\newrefformat{ass}{\savehyperref{#1}{Assumption~\ref*{#1}}}
\newrefformat{ques}{\savehyperref{#1}{Question~\ref*{#1}}}

\newrefformat{ex}{\savehyperref{#1}{Example~\ref*{#1}}}
\usepackage{bm}

\usepackage{thmtools}
\usepackage{thm-restate}
{\par\egroup\vskip 0.25ex}
\newlength\aftertitskip     \newlength\beforetitskip
\newlength\interauthorskip  \newlength\aftermaketitskip

\renewcommand{\vec}[1]{{\bm{#1}}}

\newcommand{\paren}[1]{\left(#1 \right )}

\newcommand{\Brac}[1]{\left[#1\right]}

\newcommand{\abs}[1]{\left\lvert#1\right\rvert}

\newcommand{\norm}[1]{\left\lVert#1\right\rVert}

\newcommand{\defeq}{\mathrel{\mathop:}=}%{\stackrel{\textup{def}}{=}}
\newcommand{\N}{{\mathbb N}}
\newcommand{\R}{\mathbb R}

\definecolor{DSgray}{cmyk}{0,0,0,0.7}

\newcommand{\cG}{\mathcal G}
\newcommand{\cH}{\mathcal H}
\newcommand{\cI}{\mathcal I}

\newcommand{\cL}{\mathcal L}

\newcommand{\cO}{\mathcal O}

\newcommand{\cS}{\mathcal S}
\newcommand{\cT}{\mathcal T}

\newcommand{\bigO}{O}

\DeclareMathOperator*{\argmin}{argmin} %limit displayed underneath
\DeclareMathOperator*{\argmax}{argmax} %limit displayed underneath
\usepackage{dsfont}
\usepackage{bbm}

\usepackage{subcaption}

\renewcommand{\epsilon}{\varepsilon}

\SetKwInput{KwInput}{Input}
\SetKwInput{KwOutput}{Output}
\SetKwInput{Return}{return}
\usepackage{algorithmic}

\usepackage{booktabs} % For prettier tables
\usepackage{array} % For new column types

\newcommand{\bI}{\bm{I}}

\newenvironment{hproof}{%
  \proof}{\endproof}

\newcommand{\cOT}{\textnormal{OT}}
\newcommand{\cTD}{\textnormal{TD}}
\newcommand{\cTMD}{\textnormal{TMD}}
\newcommand{\cTreeNorm}[1]{\normInline{#1}_{\textnormal{TN}_w^L}}
\newcommand{\blankTree}{T_{\bm{0}}}

\newcommand{\normInline}[1]{\Vert #1 \Vert}
\newcommand{\absInline}[1]{\lvert #1 \rvert}
\newcommand{\cOTbar}{\smash{\overline{\textnormal{OT}}}}

\title{Subsampling Graphs with GNN Performance Guarantees}

\author{
	Mika Sarkin Jain \\ Stanford University \\ \texttt{mjain4@stanford.edu} \and 
	Stefanie Jegelka \\ TUM and MIT \\ \texttt{stefje@csail.mit.edu}
    \and
	Ishani Karmarkar \\ Stanford University \\ \texttt{ishanik@stanford.edu }
  \and
  	Luana Ruiz \\ Johns Hopkins University \\ \texttt{lrubini1@jhu.edu}
    \and
	Ellen Vitercik \\ Stanford University \\ \texttt{vitercik@stanford.edu}}

\begin{document}

\maketitle

\begin{abstract}
How can we subsample graph data so that a graph neural network (GNN) trained on the subsample achieves performance comparable to training on the full dataset? This question is of fundamental interest, as smaller datasets reduce labeling costs, storage requirements, and computational resources needed for training. Selecting an effective subset is challenging: a poorly chosen subsample can severely degrade model performance, and empirically testing multiple subsets for quality obviates the benefits of subsampling. Therefore, it is critical that subsampling comes with guarantees on model performance. In this work, we introduce new subsampling methods for graph datasets that leverage the \emph{Tree Mover’s Distance} to reduce both the number of graphs and the size of individual graphs. To our knowledge, our approach is the first that is supported by rigorous theoretical guarantees: we prove that training a GNN on the subsampled data results in a bounded increase in loss compared to training on the full dataset. Unlike existing methods, our approach is both \emph{model-agnostic}, requiring minimal assumptions about the GNN architecture, and \emph{label-agnostic}, eliminating the need to label the full training set. This enables subsampling early in the model development pipeline—before data annotation, model selection, and hyperparameter tuning—reducing costs and resources needed for storage, labeling, and training. We validate our theoretical results with experiments showing that our approach outperforms existing subsampling methods across multiple datasets.
 \end{abstract}

\section{Introduction}

Graph neural networks (GNNs) have demonstrated remarkable success in graph classification tasks across diverse domains, including drug discovery \citep{gaudelet2021utilizing}, quantum chemistry \citep{stuyver2022quantum}, social networks \citep{awasthi2023gnn}, and recommendation systems \citep{sharma2024survey}. The effectiveness of GNNs critically depends on the availability of large, labeled graph datasets. However, acquiring and managing these datasets presents several key challenges:

\begin{description}[leftmargin=1.5em, labelindent=1.5em]
\item[Cost of Graph Labeling.]  Across diverse domains, labeling large graph datasets can be extremely costly. In pharmaceutical development, labeling drug molecules requires synthesizing the molecules and experimentation to characterize their properties \citep{gaudelet2021utilizing}. Meanwhile, in diagnostic settings, labeling requires expert clinical annotation \citep{li_graph_2022}. Additionally, GNNs have shown promise for combinatorial optimization, but labeling training instances requires solving large optimization problems---a computationally expensive process \citep{Cappart23:Combinatorial}.

\item[Data Storage and Streaming.] Large datasets can require substantial storage, posing challenges for data sharing, streaming, and distributed training on limited-memory devices, especially as graph datasets continue to grow \citep{zeng_gnn_2023, liu_federated_2024}. Further, many datasets contain sensitive information, such as patient records or proprietary molecular structures, making storage a potential security risk \citep{gaudelet2021utilizing, li_graph_2022}. 
    
\item[Cost of GNN Training.] Training GNNs on large datasets is computationally intensive, necessitating repeated training, architecture searches, and hyperparameter tuning. This requires significant computational resources, energy consumption, financial costs, and time-to-deployment \citep{khemani_review_2024}.  These challenges are further exacerbated when datasets exceed memory limits, requiring distributed training and additional processing overhead \citep{strubell2020energy, zeng_gnn_2023, tripathy_reducing_2020, liu_federated_2024}.

\end{description}

A natural approach to address these challenges is to subsample the graph dataset and use the subsample for GNN development. This provides a unified solution, as smaller datasets are cheaper to label, require less storage, and streamline training.  While subsampling adds a preprocessing step, it is a one-time cost amortized over multiple rounds of training and hyperparameter tuning, ultimately reducing overall computational burden. This motivates a fundamental theoretical question: \emph{how can we select a smaller set of graphs and/or nodes while preserving GNN performance?} 

This problem can be formulated as \emph{coreset construction}. A coreset is a (weighted) subset of the data that ensures a downstream task remains \emph{provably competitive} when performed on the subset instead of the full dataset \citep{bachem_practical_2017}. We aim to construct a coreset from a graph dataset such that training a GNN on the coreset achieves a loss competitive with training on the full dataset. Coreset construction is challenging as it depends both on the data and the downstream task. This is especially difficult in our case since GNNs are complex functions operating on graphs with diverse structures and feature distributions. As a result, it is difficult to formally analyze how subset selection influences model performance. Furthermore, a poorly chosen subset can severely compromise model performance. For instance, if it has graphs from only one class, the resulting GNN will fail to generalize. While it is possible to empirically assess multiple candidate subsets for quality, doing so requires labeling, storing, and training models on each subset, undermining the benefits of subsampling. This motivates the need for principled methods for coreset construction with strong guarantees on model performance.

An ideal approach should be both \emph{label-agnostic} and \emph{model-agnostic}. A coreset is \emph{label-agnostic} if its construction does not require access to graph labels and \emph{model-agnostic} if it remains representative across different GNN parameters and architectures rather than being suited for a specific model. These properties enable subsampling at the earliest stages of model development—before data annotation, model selection, and hyperparameter tuning—minimizing costs and resources needed for storage, labeling, and training.

\subsection{Our Contributions}

We present a coreset selection approach for graph classification with GNNs that, to our knowledge, is the first to come with provable guarantees on model performance. Our method is \emph{label-agnostic}, in that it does not require access to the labels of the original training set in order to construct this coreset. Our method is also \emph{model-agnostic}, in that it requires minimal assumptions on the downstream GNN architecture hyperparameters (such as layer width, type of pooling layers, and activation functions) as these details may be unknown at the time of subsampling. In contrast, prior approaches satisfy \emph{at most} one of these criteria. We make the following contributions:

\begin{description}[leftmargin=1.5em, labelindent=1.5em]
\item[Graph subsampling.] Given a training set of graphs, our method returns a subset of $k$ graphs which form a graph coreset. We provide a formal guarantee that bounds the increase in loss incurred when training a GNN on the subsampled graphs compared to the full dataset. 

\item[Node subsampling.] We extend our approach to node subsampling. We present a method that, given a graph, returns a $k$-node subgraph. Our approach again comes with strong coreset guarantees which bound the increase in loss incurred when training a GNN on these $k$-node subgraphs as opposed to the original graphs.

\item[Technical approach.] Our results are driven by new theoretical insights into the \emph{tree mover's distance (TMD)}~\citep{Chuang22:Tree}, which may be of independent interest. In particular, we introduce a linear-time algorithm to compute the TMD between a graph and any of its subgraphs, significantly improving on prior work's super-cubic runtime \citep{Chuang22:Tree}.

\item[Experiments.] We support our theoretical results with experiments comparing our method to existing subsampling approaches on datasets from OGBG and TUDatasets \citep{morris2020tudataset}. For graph subsampling, our method outperforms \citet{jin2022condensing, mirage}, and \citet{kidd}, while for node subsampling, it outperforms \citet{razin2023ability} and \citet{salha2022degeneracy}. 
\end{description}

\subsection{Related Work}\label{sec:related-work}

\paragraph{Coreset construction.}
Coresets have been extensively studied in classical ML for Euclidean data, with applications in mean approximation, regression, clustering, and convex optimization \citep{bachem_practical_2017, mirzasoleiman2020coresets, woodruff_coresets_2024, cohen-addad_improved_2022, tukan_coresets_2020}. These methods construct small, weighted subsets of datapoints that provably approximate solutions obtained on the full dataset. More recently, coresets have been explored for improving the efficiency of training deep learning models \citep{yang_towards_2023, ding_spectral_2024, mirzasoleiman2020coresets}.  

Despite this progress, coreset selection for structured data, particularly graphs, remains largely unexplored. \citet{ding_spectral_2024} propose a subset selection approach for node classification, but it lacks the provable guarantees typical of coreset methods and differs fundamentally from our work, which focuses on graph classification. To our knowledge, ours is the first work to introduce coreset selection for graph classification and to provide formal guarantees on GNN performance when trained on the coreset.

\paragraph{Graph dataset compression.}{ A related but distinct line of research focuses on graph condensation, which reduces the size of a GNN training dataset by building a smaller, \emph{synthetic} dataset that resembles the original. Most works in this area target node-level task, making them unsuitable for comparison with our graph-level approach. For graph-level tasks, DosCond \citep{jin2022condensing} and KiDD \citep{kidd} use \emph{gradient matching}, which requires access to the full GNN hyperparameter specifications and the training trajectories of a GNN trained on the original dataset. These methods build the synthetic graphs to replicate the training trajectories on the full dataset. Thus, they are neither label- nor model-agnostic and lack theoretical guarantees. Moreover, gradient matching requires training a model on the full dataset to construct the subsample, which undermines key practical motivations of subsampling.
}

\paragraph{Model-agnostic graph dataset compression.} \citet{mirage} propose a model-agnostic graph dataset compression procedure, MIRAGE, that uses the training graphs' computation trees to compress the training set. Unlike gradient-matching approaches, MIRAGE is model-agnostic: it does not require knowledge of downstream model parameters to build the subsample. However, it is \emph{not} label-agnostic, as its sampling procedure relies on labels from the full dataset.

\paragraph{Other applications of TMD.} \citet{georgiev23} uses a normalized variant of TMD to select training sets for neural algorithmic reasoning tasks. However, their work is not directly applicable to graph or node subsampling.
\section{Notation and Background}\label{sec:prelim}

\paragraph{Notation.} An undirected graph $G = (V, E, f)$ consists of a node set $V$, an edge set $E$, and node features $f \in \mathbb{R}^{p \times |V|}$. Each node $v \in V$ has an associated feature vector $f^v \in \mathbb{R}^p$. The neighborhood of a node $v \in V$ is denoted by $N_G(v)$. We use $\mathcal{G}$ to represent a set of graphs. An $r$-rooted tree is denoted as $T = (V, E, f, r)$. The all-ones vector, all-zeros vector, and identity matrix are represented by $\bm{1}$, $\bm{0}$, and $\bm{I}$, respectively. The notation $\normInline{\cdot}$ is used to denote an arbitrary norm. Visualizations of definitions are provided in Appendix~\ref{sec:notation}.

\paragraph{Message-Passing Graph Neural Networks (GNNs).} We describe the \emph{Graph Isomorphism Network (GIN)} \citep{xu19gin} as a representative example of a GNN architecture, though the results presented extend to other architectures. A GIN is parameterized by $L$ functions $\phi^{(\ell)}: \mathbb{R}^d \to \mathbb{R}^d$, typically implemented as shallow feed-forward neural networks, and a fixed weight $\eta > 0$. Each node is initialized with an embedding $z_v^{(0)} = f^v$ and updates its embedding as follows:
\begin{align*}
    &\text{\textbf{Message passing:} for } \ell \in [L-1], \quad
    z_v^{(\ell)} = \phi^{(\ell)} \Big(z_v^{(\ell-1)} + \eta \sum\nolimits_{u \in N(v)} z_u^{(\ell-1)}\Big), \\
    &\text{\textbf{Graph readout:} } \quad
    h(G) = \phi^{(L)} \Big(\sum\nolimits_{v \in V} z_v^{(L-1)}\Big).
\end{align*}
Other GNN architectures may generalize this framework by replacing the sum operator with an alternative aggregation function, among other variations.

\paragraph{Tree Mover's Distance (TMD).}  
TMD \citep{Chuang22:Tree} quantifies graph similarity by representing each graph as a multiset of \emph{computation trees}. A node's computation tree is constructed by first including its immediate neighbors at the first level, followed by their neighbors at the next level, continuing recursively up to a predefined depth. TMD then measures the similarity between graphs by comparing their multisets of computation trees using optimal transport (OT).

\begin{figure}[t]
  \centering
  \includegraphics[width=.4\textwidth]{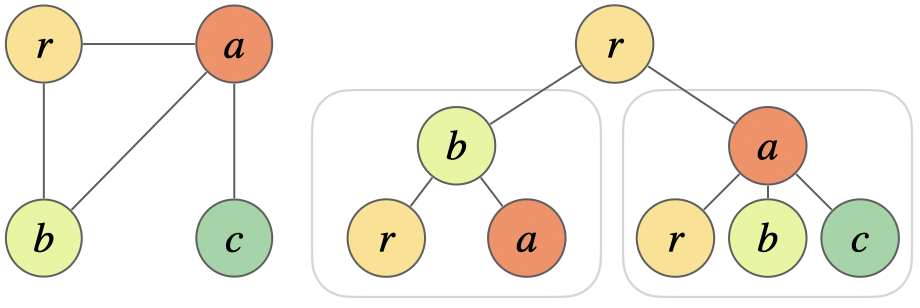}
  \caption{A graph $G$ and the depth-2 computation tree $T = T^2_r(G)$ of node $r$ (Definition~\ref{def:tree}). The set $\cT_r(T)$ (Definition~\ref{def:multiset}) consists of the two boxed subtrees.}
  \label{fig:tree}
\end{figure}

\begin{definition}[Computation tree]\label{def:tree} 
Given $G = (V, E, f)$ and a node $v \in V$, let $T_v^1(G) = (v, \emptyset, \{f^v\}, v)$ be a singleton tree rooted at $v$, containing only the node $v$ (with node feature vector $f^v$) and
no edges. Inductively, the depth-$L$ computation tree $T_v^L(G)$ is formed by attaching the neighbors of each leaf node in $T_v^{L-1}(G)$. Concretely, for each leaf $\ell$ of $T_v^{L-1}(G)$ and each neighbor $u \in N_G(\ell)$, we add a new node $\ell_u$ and connect $\ell$ to $\ell_u.$ The multiset of depth-$L$ computation trees defined by {$G$} is $\cT_G^L \defeq \{T_v^{L}(G)\}_{v \in V}$. An example is depicted in Figure~\ref{fig:tree}. 
\end{definition}

\begin{definition}[Tree multisets]\label{def:multiset} Let $T = (V, E, f, r)$ be an $r$-rooted tree of depth $\ell$. $\cT_r(T)$ is the multiset of depth $(\ell-1)$ computation trees rooted at each neighbor {$u \in N_T(r)$}. That is, each $u \in N_T(r)$ has subtree $T_u$ of depth $\ell-1$, and $\cT_r(T) \defeq \cup_{u \in N_T(r)} {T^{(\ell-1)}_u}({T_u})$. (Here $\cup$ is multiset union).
\end{definition}

\begin{definition}[Optimal transport]\label{def:OT} Let $X = \{x_{i}\}_{i=1}^q$ and $Y = \{y_{i}\}_{i=1}^q$ be two multisets. Given a metric $d: X \times Y \to \R_{+}$, let $C\in \R^{q\times q}$ be a matrix where $C_{i,j} = d(x_i, y_j)$ is the \emph{transportation cost} between $x_i$ and $y_j$. A \emph{transportation plan} $\gamma$ is a $q \times q$ nonnegative doubly-stochastic matrix, i.e., whose rows and columns sum to 1. The cost of a plan $\gamma$ is $\langle \gamma, C \rangle \defeq \sum_{i,j} \gamma_{i,j} C_{i,j}.$ The (unnormalized) Wasserstein distance $\cOT_d$ is defined as ${\cOT_d (X, Y) \defeq q \min_{\gamma \in \Gamma(X, Y)} \langle C, \gamma \rangle}$, where ${\Gamma(X, Y) \defeq \{\gamma \in \R^{q \times q}_+ \mid \gamma \bm{1} = \gamma^\top \bm{1} = \bm{1} \}}$ is the feasible set of transportation plans. A plan $\gamma^*$ achieving the minimum is called an \emph{optimal transport plan.}
\end{definition} 

To compare tree multisets with OT requires a metric between trees. To achieve this, TMD employs the \emph{tree distance}, which measures similarity by comparing the features of the root nodes and \emph{recursively} comparing their subtrees. Specifically, for each rooted tree, the collection of subtrees rooted at each neighbor of the root is constructed, as illustrated in Figure~\ref{fig:tree}. Since OT operates on multisets of equal cardinality, comparability is ensured by padding the smaller set with ``blank trees'' (isolated nodes with a feature vector of $\vec{0}$) using a padding function $\rho(\cdot, \cdot)$.   

\begin{definition}[Tree distance]\label{def:TD} Let $T_a = (V_a, E_a, f_a, r_a)$ and $T_b = (V_b, E_b, f_b, r_b)$ be two rooted trees with maximum depth $\ell$  and let $w: \N \to \R_{+}$ assign a weight to each depth level.
   The \emph{tree distance} is defined recursively as:
\begin{align*}
    \cTD_w(T_a, T_b) \defeq \norm{f_a^{r_a} - f_b^{r_b}} + \begin{cases}
        w(\ell) \cdot \cOT_{\cTD_w}(\rho(\cT_{r_a}(T_a), \cT_{r_b}(T_b)))  &\text{if }\ell > 1 \\
        0 &\text{otherwise.} 
    \end{cases}\end{align*}
\end{definition}

The tree distance $\cTD_w$ provides a way to compare individual computation trees by aligning their root features and recursively matching their subtrees using optimal transport. Extending this to entire graphs, TMD represents each graph as a multiset of computation trees and defines the distance between two graphs as the optimal transport cost between these multisets, using the tree distance between the computation trees in the multisets.  For brevity, we use the notation $\cOTbar(\cdot, \cdot) = \cOT_{\cTD_w}(\rho(\cdot, \cdot))$ to denote the optimal transport cost with tree distance and padding.

\begin{definition}[Tree mover's distance]\label{def:TMD} Given graphs $G$ and $G'$, weight function $w: \N \to \R$, and depth parameter $L > 0$, we define 
$\cTMD_w^L(G, G') \defeq \cOTbar(\cT_G^L, \cT_{G'}^L).$
We also define the \emph{tree norm} of a graph $G$ as $\cTreeNorm{G} \defeq \cTMD_w^L(G, \emptyset)$, where $\emptyset$ represents an empty graph.
\end{definition}
By decomposing graphs into local computation trees and comparing them globally, TMD captures both structural and feature-based similarities at multiple scales. As a pseudometric, it satisfies non-negativity and symmetry.

\newcommand{\xmark}{\ensuremath{\boldsymbol{\times}}}

\section{Graph Subsampling}\label{sec:graph_drop}

Here we present our approach for subsampling graphs in a dataset while preserving the performance of a downstream GNN.  Our method relies on the following Lipschitz bound that relates a GNN's stability to TMD\footnote{While Theorem 8 in \citet{Chuang22:Tree} is stated for GINs, they extend it to other GNN architectures.}.

\begin{theorem}[Informal, Theorem 8 by \citet{Chuang22:Tree}, restated]\label{thm:stable} {There exists a weight function $w$ such that} for any $(L-1)$-layer message-passing GNN with readout $h: \cG \mapsto \R^d$ and layer-wise Lipschitz constants $\phi_\ell$, the following holds for any two graphs $G_a, G_b$: \[\norm{h(G_a) - h(G_b)}\leq \cTMD_{w}^{L} (G_a, G_b) \cdot \prod_{\ell=1}^{L-1} \phi_\ell.\] 
\end{theorem}

Let $X= \{G_1, ..., G_n\}$ be our graph dataset. Given a budget $k$, we aim to choose a subset $\cI \subset [n]$ of size $k$ such that a GNN trained on the subsampled set {$\{G_i\}_{i \in \cI}$} obtains similar readouts and loss as if it were trained on the entire set $X$.
To ensure $\cI$ is representative of $X$, we select $\cI$ by optimizing a medoids objective, which quantifies how well each graph $G_i \in X$ is represented by at least one graph in $\cI$. 

\begin{definition}[Medoids objective] Let $X= \{G_1, ..., G_n\}$ be a graph dataset and $\cI \subset [n]$ with $\absInline{\cI} \leq k$. The medoids objective value of $\cI$ with respect to distance $D : X \times X \to \R_{\geq 0}$ is defined by 
\begin{align*}
    f_D(\cI; X) = \frac{1}{\abs{X}}\sum\nolimits_{i \in [n]} \min_{j \in\cI} D(G_i, G_j).
\end{align*}
For $j \in \cI$, let $\tau_j$ be the number of graphs closest to $G_j$:
\[\tau_j \defeq \abs{\{i \in [n]: D(G_i, G_j) < D(G_i, G_k),~\forall k \neq j\}},\]
breaking ties arbitrarily. We call $\tau_j$ the size of cluster $j$. 
\end{definition}

The provide a lemma that bounds the difference in a GNN's loss on the (weighted) subsampled dataset $\cI \subset X$ and the full dataset $X$ in terms of the medoids objective with respect to TMD. Proofs of results in this section are in Appendix~\ref{sec:apx_graph_drop}.

\begin{restatable}{lemma}{tmdgen}\label{lemma:tmd-gen} Let $\cH$ be a hypothesis class of $(L-1)$-layer GNNs $h: \cG \to \R^d$, where the $\ell$-th layer has Lipschitz constant at most $\Phi_\ell$. Let $\cL: \R^d \to \R$ be an $M$-Lipschitz loss function, let {$y = (y_1, ..., y_n)$ be labels for $X$}, and $\cI$ be a subset of $[n]$. For any GNN $h \in \cH$ and graph $G \in \cG$,  
\begin{align}
    \Big\lvert{\frac{1}{n} \sum\nolimits_{i \in \cI} \tau_{i} \cL(h(G_i); y_i) - \frac{1}{n}\sum\nolimits_{i \in [n]}\cL(h(G_i); y_i)}\Big\rvert \nonumber \leq  M \cdot f_{\cTMD_w^L}(\cI; X) \cdot {\prod\nolimits_{\ell \in [L-1]} \Phi_\ell} . \label{eq:mediods_bound}
\end{align}
\end{restatable}
\begin{hproof} We use the Lipschitz constant $M$ of $\cL$ and Theorem~\ref{thm:stable} to bound the average deviation in the loss on $X$ from the loss on $\cI$ (reweighted according to the $\tau_i$'s). 
\end{hproof}

If $f_{\cTMD_w^L}(\cI; X)$ is small, then each $G_i \in X$ is close to some subsampled graph, which keeps the overall loss on the subsampled set similar to the loss on $X$. If we minimize loss over $\cI$, the next corollary shows that the resulting hypothesis will incur only a small increase in loss on $X$.
\begin{restatable}{corollary}{erm}\label{cor:ERM}
    Suppose $M \cdot \prod_{\ell \in [L-1]} \Phi_\ell \leq c$, $f_{\cTMD_w^L}(\cI; X) \leq \epsilon$, and $\hat{h} \in \cH$ minimizes the weighted loss $\sum_{i \in \cI} \tau_{i} \cL(h(G_i); y_i).$ Then $
    \frac{1}{n}\sum\nolimits_{i \in [n]}\cL\big(\hat{h}(G_i); y_i\big) \leq \min_{h \in \cH}\frac{1}{n}\sum\nolimits_{i \in [n]}\cL(h(G_i); y_i) + 2c\epsilon.$
\end{restatable}
% \paragraph{Graph subsampling.}
% \begin{figure}[t]
% \centering
%     \includegraphics[width=\textwidth]{figures/GraphSubsampling.png}
% \caption{Test accuracy versus fraction of graphs in the training set, subsampled with our approach and existing model-agnostic methods.}
% \label{fig:graph_subsampling}
% \end{figure}
%Remark~\ref{remark:tmd-gen-remark}
Hence, the additional training loss incurred by training on $\cI$ instead of $X$ is bounded by an error proportional to $f_{\cTMD_w^L}(\cI; X).$ This bound could be combined with prior work on the VC dimension of GNNs~\citep[e.g.,][]{scarselli2018vapnik} to obtain bounds on the \emph{test loss} as well.

Corollary~\ref{cor:ERM} motivates selecting $\cI$ to minimize $f_{\cTMD_w^L}(\cI; X).$
% \begin{definition}[Graph subsampling problem]\label{def:graph-subsampling} Let $X= \{G_1, ..., G_n\}$ be a set of graphs and $k$ be a budget. The \emph{graph subsampling problem} is to select a subset $\cI \subset [n]$ of $k$ graphs that minimizes the objective $f_{\cTMD_w^L}(\cI; X).$
% \end{definition}
Importantly, this optimization problem is independent of graph labels. Training a GNN on the subsampled graphs in only requires knowing the labels of those graphs. Additionally, our results make mild assumptions on the GIN architecture---we need only know its depth $L$. Though the $k$-medoids problem---and thus optimizing $f_{\cTMD_w^L}(\cI; X)$---is NP-hard~\citep{kazakovtsev2020application}, {there are} efficient approximation algorithms, for example Python's $k$-medoids algorithm from the sklearn package \citep{sklearn_api}, which we use in our experiments.

\paragraph{Other pseudo-metrics.} Given these results, a natural question is whether the TMD pseudo-metric is essential for the stability result in Theorem~\ref{thm:stable}. One might hope that other graph pseudo-metrics could also be used to bound GNN stability, thereby yielding analogs of Corollary~\ref{cor:ERM}.  We express this intuition as the following conjecture.

\begin{conjecture}\label{conj:TMD-not-needed} Let $D$ denote a graph pseudo-metric. There exists a function $C : \{\phi_\ell\}_{\ell=1}^{L-1} \to \R_{> 0}$ such that for any $(L-1)$-layer message-passing GNN with readout function $h: \cG \to \R^d$ and layer-wise Lipschitz constants $\phi_\ell$, the following holds for any two graphs $G_a, G_b$: $\norm{h(G_a) - h(G_b)} \leq C\paren{\phi_1, ..., \phi_{L-1}} \cdot D(G_a, G_b)$. 
\end{conjecture}

Surprisingly, we show that this conjecture is \emph{provably} false for four of the most widely used graph pseudo-metrics: Weisfeiler–Lehman (WL) \citep{shervashidze2011weisfeiler}, WL-Optimal Transport (WL-OA) \citep{kriege2016valid}, shortest-paths (SP) \citep{borgwardt2005shortest}, and graphlet sampling (GS) \citep{shervashidze2009efficient}. 

\begin{restatable}{theorem}{tmdneeded}\label{thm:TMD-is-needed}
    Let $D^L$ denote any of the following pseudo-metrics: GS, $L$-layer WL, $L$-layer WL-OA, or the $L$-layer SP. Then Conjecture~\ref{conj:TMD-not-needed} is false for $D = D^L$.
\end{restatable}

Consequently, our theoretical findings and the main result in Corollary~\ref{cor:ERM} rely \emph{crucially} on TMD as the underlying pseudo-metric. See Appendix~\ref{sec:thm-tmd-is-needed} for details.
\section{Node Subsampling}\label{sec:node_drop}

In this section, we present our approach for subsampling nodes of a graph dataset so that the performance of a downstream GNN trained on the subsample is preserved. Proofs for results in this section are in Appendix~\ref{sec:apx-node}.

Suppose we have a graph dataset $X= \{G_1, ..., G_n\}$ where $G_i = (V_i, E_i)$, and a node budget $k$. For a subset $S_i \subseteq V_i$ of $k$ nodes, let $G_i[S_i]$ denote the subgraph of $G_i$ induced by $S_i.$ Our goal is to {select these subsets so} that a GNN produces similar readouts on the original dataset $X$ and on the induced subgraphs $X' = \{G_1[S_1], ..., G_n[S_n]\}$. The following corollary shows that if the GNN layers have small Lipschitz constants and the distances $\cTMD(G_i[S_i], G_i)$ are small, then training over $X'$ yields a nearly optimal predictor over $X$. 

\begin{restatable}{corollary}{corrermfirst}\label{corr:erm-first} Let $\cH$ be a set of $(L-1)$-layer GNNs $h: \cG \to \R^d$, where the $\ell$-th layer has Lipschitz constant at most $\Phi_\ell$. Let $\cL: \R^d \to \R$ be an $M$-Lipschitz loss function. 

Suppose that $M \cdot \prod_{\ell \in [L-1]} \Phi_\ell \leq c$ and $\cTMD_w^L(G_i, G_i[S_i]) \leq \epsilon_i$ for all $i \in [n]$. Finally, let $\hat{h} \in \cH$ be a GNN with minimum loss over the subsampled training set with respect to the training labels $y_1, \dots, y_n$: $\hat{h} = \argmin\nolimits_{h \in \cH}\sum\nolimits_{i \in [n]} \cL(h(G_i[S_i]); y_i).$
Then $\hat{h}$ has near-optimal loss over the original training set:
\begin{align*}
    \frac{1}{n}\sum\nolimits_{i \in [n]} \cL\Big({\hat{h}(G_i); y_i}\Big) \leq  \min_{h \in \cH} \frac{1}{n}\sum\nolimits_{i \in [n]} \cL(h(G_i); y_i) + \frac{2c}{n}\sum\nolimits_{i \in [n]} \epsilon_i.
\end{align*}
\end{restatable}

Thus, the additional {loss incurred by} training on $X'$ rather than $X$ is proportional to the average of {$\cTMD_{w}^L(G_i, G_i[S_i])$}, so for each $G \in X$, we would ideally solve:
\begin{equation}
    \min_{S \subset V: \abs{S} \leq k} \cTMD_w^L(G, G[S]).\label{eq:node-subsampling}
\end{equation}

We face {two} key challenges in doing so. First, the number of candidate subsets $\{S \subset V: \abs{S} \leq {k}\}$ grows exponentially.  Indeed, solving \eqref{eq:node-subsampling} is NP-hard (see Appendix~\ref{sec:hardness}).
 We therefore 
 restrict
 our search to an appropriately chosen feasible set $\cS$.
In our experiments, we combine well-motivated, fast heuristics for selecting small candidate sets $\cS$ based on prior analyses \citep{salha2022degeneracy, razin2023ability, alimohammadi2023local}. (Details are in Appendix~\ref{sec:heuristic}.)

The next challenge is that computing $\cTMD_w^L(G, G[S])$ for each $S \in \cS$ can be expensive: the algorithm by \citet{Chuang22:Tree} takes $\cO(L|V|^4)$ time. To address this, we prove that, surprisingly, computing $\cTMD_w^L(G, G[S])$ is equivalent to a much simpler optimization problem. 

\begin{restatable}{theorem}{nodesamplingrelaxed}\label{thm:node-subsampling-relaxed} Let $G = (V, E, f)$ be a graph and $\cS$ be a set of candidate node subsets. Then \begin{align}\label{eq:final-opt-problem}
    \argmin_{S \in \cS} \cTMD_w^L(G, G[S]) = \argmax_{S \in \cS} \cTreeNorm{G[S]}.
\end{align}
\end{restatable}

We prove Theorem~\ref{thm:node-subsampling-relaxed} in Section~\ref{sec:proof} and in Section~\ref{sec:alg}, we provide a linear-time algorithm for computing tree norms and, thereby, the solution to Equation~\eqref{eq:final-opt-problem}.

\begin{restatable}{theorem}{algorithm}\label{thm:node-subsampling-relaxed-algorithm} Given a graph $G = (V, E, f)$, Algorithm~\ref{alg:1} computes $\cTreeNorm{G}$ in $\bigO(\abs{E}L)$ time.
\end{restatable}

\subsection{TMD Between Graphs and Subgraphs}\label{sec:proof}

In this section, we sketch our proof of Theorem~\ref{thm:node-subsampling-relaxed}. In doing so, we prove several properties of the TMD which may be of independent interest given the broad applications of TMD to out-of-distribution generalization. Our analysis crucially relies on the following lemma, which shows that an appealing simple \emph{identity} transport plan can be used to compute $\cTMD$ between a graph and its induced subgraphs.

\begin{restatable}{lemma}{generaldecomposition}\label{lemma:general-decomposition} Let {$w: \N \to \R$ be a weight function and} $G = (V, E, f)$ be a graph.
For $S \subseteq V$, define the identity transportation plan $\bm{I}$ that maps $T_v(G)$ to $T_v(G[S])$ if $v \in S$ and to $\emptyset$ otherwise. Then $\bm{I}$ is an optimal transport plan for
\begin{align}\label{eq:ot-problem}
    \cTMD_w^L(G, G[S]) := \cOTbar\paren{\cT_G^L, \cT_{G[S]}^L}.
\end{align}
Consequently, we can decompose $\cTMD_w^L(G, G[S])$ as: 
\begin{align}
    \cTMD_w^L(G, G[S]) \label{eq:decomposition} &= \underbrace{\sum_{v \notin S} \|f^v\|}_{\textnormal{Deleted nodes' features}} +\underbrace{w(L-1)\sum_{v \notin S} \cOTbar
    \left(\cT_v\left(T_v^L(G)\right),\,\emptyset\right)}_{\textnormal{Cost of removing deleted nodes' trees}}\\
    &+\;
    \underbrace{w(L-1)\sum_{v \in S} \cOTbar
    \left(\cT_v\left(T_v^L(G)\right),\, \cT_v\left(T_v^L(G[S])\right)\right)}_{\textnormal{Cost of matching retained nodes' trees}}.\nonumber
\end{align}
\end{restatable}
\begin{hproof} We prove the statement by induction on $L$. In the base case $(L= 1)$, by definition, $\cOT(\cT_G^1, \cT_{G[S]}^1)$ 
equals the sum of features of the nodes that are in $G$ but not in $G[S]$, i.e., $\cOT(\cT_G^1, \cT_{G[S]}^1) = \sum_{v \notin S} \Vert f^v \Vert = \langle C, \bm{I} \rangle$, so the base case holds. Inductively, we use the TD's recursive formulation to reduce OT problems of depth $L$ to those of depth $L-1$.
\end{hproof}

Next, we use Lemma~\ref{lemma:general-decomposition} to characterize how TMD accumulates as we remove a sequence of nodes. Lemma~\ref{lemma:chain-simple} shows that this accumulation is \emph{additive} under certain conditions, which is unexpected given the TMD's combinatorial nature.

\begin{restatable}{lemma}{finegraineddecomp}\label{lemma:chain-simple} 
For $T \subset S \subset V$, $\cTMD_w^L(G, G[S \setminus T ]) = \cTMD_w^L(G, G[S]) + \cTMD_w^L(G[S], G[S\setminus T])$.  
\end{restatable}

 This equality is noteworthy because the triangle inequality only guarantees $\cTMD_w^L(G, G[S \setminus T ]) \leq \cTMD_w^L(G, G[S]) + \cTMD_w^L(G[S], G[S\setminus T])$. However, we show that the triangle inequality is \emph{always} tight when $T \subset S \subset V$.

\begin{hproof}[Proof sketch of Lemma~\ref{lemma:chain-simple}] We apply Lemma~\ref{lemma:general-decomposition} {inductively over $L$}. When $L=1$, the second and third summands of \eqref{eq:decomposition} equal 0: $\cT_G^1$ and $\cT_{G[S]}^1$ are multisets of depth 1, so for any $v \in V$, $T_v^1(G) = \{v\}$ and thus $\cT_v(T_v^1(G)) = \emptyset$. Likewise, for any $v \in S$, $\cT_v(T_v^1(G[S])) = \emptyset$. Therefore, 
\begin{align*}
    \cTMD_w^1(G, G[S \setminus T]) = \sum\nolimits_{v \notin (S \setminus T) } \norm{f^v} = \sum\nolimits_{v\notin S} \norm{f^v} + \sum\nolimits_{v \in T} \norm{f^v}, 
\end{align*} 
because for $T \subset S$, $\{v \in V : v \notin (S \setminus T) \} = \{v \in V : v \notin S\} \sqcup T$, where $\sqcup$ denotes the \emph{disjoint union}.
Similarly, 
\begin{align*}
    \cTMD_w^1(G, G[S]) = \sum\nolimits_{v \notin S} \norm{f^v} \text{\,\,\,\,and\,\,\,\,} \cTMD_w^1(G[S], G[S \setminus T]) = \sum\nolimits_{v \in T} \norm{f^v}, 
\end{align*} 
thus verifying the base case. The intuition is similar for $L>1$ but requires
care to handle the recursive OT terms.
\end{hproof}

Finally, we can prove Theorem~\ref{thm:node-subsampling-relaxed}.

\begin{proof}[Proof of Theorem~\ref{thm:node-subsampling-relaxed}] Let $S \in \cS$. By Lemma~\ref{lemma:chain-simple} with $T = S$, $\cTreeNorm{G} =  \cTMD_w^L(G, \emptyset) = \cTMD_w^L(G, G[S]) + \cTMD_w^L(G[S], \emptyset) = \cTMD_w^L(G, G[S]) + \cTreeNorm{G[S]}.$  Thus, 
\begin{align*}
    \max_{\substack{S \in \cS \\ \abs{S} = k}}
 \cTreeNorm{G[S]} \equiv \max_{\substack{S \in \cS \\ \abs{S} = k}} \cTreeNorm{G} - \cTMD_w^L(G, G[S]) \equiv \min_{\substack{S \in \cS \\ \abs{S} = k}} \cTMD_w^L(G, G[S]). \tag*{\qedhere} 
\end{align*}
\end{proof}

\subsection{Faster Algorithm for Tree Norms}\label{sec:alg}
We next leverage Lemma~\ref{lemma:general-decomposition} to obtain  Algorithm~\ref{alg:1}, a faster algorithm for computing $\cTreeNorm{G}$ for $G = (V, E, f).$ The algorithm by \citet{Chuang22:Tree} requires $\bigO(L \abs{V}^4)$ time, whereas ours {has runtime} $\bigO(L |E|)$. Algorithm~\ref{alg:1} is based on the fact that by Lemma~\ref{lemma:general-decomposition}, $\cTreeNorm{G}$ is essentially a weighted sum of the number of vertices in all of its depth-$L$ computation trees, which Algorithm~\ref{alg:1} {computes}. Its runtime is dominated by the cost of $L$ matrix-vector {multiplies} with the adjacency matrix, which takes $\bigO(|E|)$-time. Proofs of correctness and runtime are in Appendix~\ref{sec:apx-node}. \\

\RestyleAlgo{ruled}
\SetKwComment{Comment}{/* }{ */}
\begin{algorithm2e}[H]
\caption{\text{TreeNorm}($G, L, w$)}\label{alg:1}
\KwInput{Graph $G = (V, E, f)$ {with adjacency matrix $A$}, weights $w: \{1, ..., L-1\} \to \R_+, L\geq 1$}
Define $x \in \R^{|V|}$ such that $x_v = \norm{f^v}$ \label{line:x}\; 
Initialize $z^{(0)} = x$\; 
\For{$\ell \in [{L-1}]$}{
    $z^{(\ell)} \gets A z^{(\ell-1)}$\label{line:z}\;
}
$b \gets z^{(0)} + \sum_{\ell = 1}^{L-1} \paren{\prod_{t=1}^\ell w(L-t)} \cdot z^{(\ell)}$\;  
\Return{$\normInline{b}_1$}
\end{algorithm2e}

\section{Experiments}\label{sec:experiments}

\paragraph{Graph subsampling.} We compare our approach with several graph subsampling and condensation methods. KiDD \citep{kidd} and DosCond \citep{jin2022condensing} condense datasets into small synthetic graphs but are not label-agnostic. For a fair comparison, we modify them to operate in a label-agnostic manner by removing their label-aware components in the ``without labels'' section of Tables~\ref{tab:all-results-graph-subsampling1} and \ref{tab:all-results-graph-subsampling2}, as detailed in Appendix~\ref{sec:additional-background-experiments-vallabel}. Since these methods do not support randomized train/test splits and rely on fixed initializations, we do not randomize over splits but do randomize over GNN parameter initializations. MIRAGE \citep{mirage} is another condensation method that subsamples computation trees, but persistent runtime errors in its implementation prevented thorough benchmarking, an issue also reported by \citet{sun2024gc}. We successfully ran MIRAGE on two datasets for certain sampling percentages, as detailed in Appendix~\ref{sec:additional-background-experiments-mirage}.  In addition to these methods, we construct three additional baselines. \emph{WL} applies $k$-medoids clustering using the widely-used Weisfeiler-Lehman (WL) kernel \citep{shervashidze2011weisfeiler}. \emph{Random} performs uniform graph subsampling. \emph{Feature} applies $k$-medoids clustering based only on node features, ignoring graph structure (see Appendix~\ref{sec:additional-background-experiments-feature}). This baseline is particularly useful for evaluating the impact of structure-aware methods such as TMD.  We use an 80/20 train/test split and select between 1\% and 10\% of the training graphs. A GNN is trained on the selected graphs, and we report the average test performance with 95\% confidence intervals over 20 trials with random train/test splits and network initializations.  

Our findings are summarized in Tables~\ref{tab:all-results-graph-subsampling1} and \ref{tab:all-results-graph-subsampling2}. Following \citet{jin2022condensing}, we measure test AUC-ROC on OGBG datasets and classification test accuracy on the remaining datasets. ``Wins'' count how often a method outperforms others, while ``Fails'' count how often a method falls below 50\% accuracy, as all tasks involve binary classification.  TMD consistently ranks first or second across nearly all datasets and subsampling percentages, except on NCI1, where no method outperforms Random. While Random achieves strong performance on NCI1, it performs poorly on most other datasets, a trend also observed with DosCond, which achieves strong results on OGBG-MOLBACE but struggles without labels. Overall, TMD attains the highest total wins across datasets, as shown in Table~\ref{tab:wins-fails-graph}. Additionally, while DosCond and KiDD exhibit high variance, particularly on PROTEINS, TMD maintains stable performance across different subsampling percentages. This consistency is reflected in its low fail count in Table~\ref{tab:wins-fails-graph}, underscoring the robustness of our approach.  

\begin{table}[t]
\caption{Summary of graph and node subsampling performance. The ``Wins" column counts the number of times each method achieves the best test performance across datasets and sampling percentages. The ``Fails" column counts instances where test performance falls below 50\% (worse than random chance). No fails were observed for node subsampling. TMD achieves the highest number of wins across both tasks, while the strong performance of Random is primarily observed on the NCI1 dataset. Full results are presented in Tables~\ref{tab:all-results-graph-subsampling1}, \ref{tab:all-results-graph-subsampling2}, and \ref{tab:combined-results-nodes}.}
\centering
\label{tab:summary}
\begin{tabular}{cc} 
\begin{subtable}{0.48\textwidth}
    \centering
    \caption{Graph subsampling}
    \label{tab:wins-fails-graph}
    \begin{tabular}{l | c | c}
    \hline
    \textbf{Method} & \textbf{Wins} $\uparrow$ & \textbf{Fails} $\downarrow$ \\
    \hline
    DosCond & 11 & 20 \\
    KiDD & 7  & 7 \\
    Feature & 13  & 3 \\
    WL & 8 & 4 \\
    Random & 18 & 7 \\
    \textbf{TMD} & \textbf{23} & \textbf{2} \\
    \hline
    \end{tabular}
\end{subtable}
\begin{subtable}{0.48\textwidth}
    \centering
    \caption{Node subsampling}
    \label{tab:wins-fails-node}
    \begin{tabular}{l | c }
    \hline
    \textbf{Method} & \textbf{Wins} $\uparrow$  \\
    \hline
    RW & 15 \\
    k-cores & 5  \\
    Random & 16 \\
    \textbf{TMD} & \textbf{25} \\
    \hline
    \end{tabular}
\end{subtable}
\end{tabular}
\end{table}

\paragraph{Node subsampling.} To our knowledge, no existing methods focus on node subsampling for graph classification, so we adapt two related approaches as benchmarks. \emph{K-cores}, proposed by \citet{razin2023ability}, is a node selection heuristic based on $k$-core decomposition, which identifies structurally important nodes by iteratively pruning low-degree nodes. \emph{RW}, introduced by \citet{salha2022degeneracy}, is a random-walk-based heuristic originally designed for subgraph sampling in large graph autoencoders. Additionally, we compare against \emph{Random}, which performs uniform node subsampling.  For our proposed node subsampling method (Theorems~\ref{thm:node-subsampling-relaxed} and~\ref{thm:node-subsampling-relaxed-algorithm}), we construct the candidate set $\mathcal{S}$ using breadth-first search (BFS) trees, leveraging the fact that BFS preserves critical motifs in real-world networks~\citep{alimohammadi2023local}. We further augment $\mathcal{S}$ with subsets generated by the RW and k-cores heuristics. The final subset is then selected using TMD. Additional details are provided in Appendix~\ref{sec:heuristic}.  Datasets with relatively larger graphs, such as COX2, PROTEINS, and DD, are well-suited for node subsampling, whereas MUTAG, NCI1, OGBG-MOLBACE, and OGBG-MOLBBBP contain fewer than 35 nodes per graph. For completeness, we include results on MUTAG to demonstrate that our method remains competitive even on smaller graphs.  

Table~\ref{tab:combined-results-nodes} compares test accuracy across these methods for sampling fractions ranging from 10\% to 90\%. We report averages with 95\% confidence intervals over 20 trials with random neural network initializations and train/test splits. The ``W'' row counts the number of times a method outperforms others. TMD consistently ranks among the top two methods and achieves the best overall performance (Table~\ref{tab:wins-fails-node}).  \\

%\newpage
\begin{table}[H]
\centering
\caption{Graph subsampling performance across sampling percentages for TUDatasets, reported as mean ± confidence bar of accuracy (ACC) or area under the ROC (ROC-AUC). Best and second-best per column are in dark/light green, respectively. W is total wins per method.}
\vspace{.1in}
\label{tab:all-results-graph-subsampling1} 
\begin{subtable}{\textwidth}
\captionsetup[subtable]{aboveskip=0pt, belowskip=0pt}
\centering
\resizebox{1.0\textwidth}{!}{%
\begin{tabular}{l | c | c | c | c | c | c | c | c | c | c | c | c}
 \hline
 \textbf{Method} & \textbf{1\%} & \textbf{2\%} & \textbf{3\%} & \textbf{4\%} & \textbf{5\%} & \textbf{6\%} & \textbf{7\%} & \textbf{8\%} & \textbf{9\%} & \textbf{10\%} & \textbf{W} & \textbf{F} \\
\hline
\multicolumn{12}{c}{\textbf{With labels}} \\
\hline
\textbf{Doscond} & 0.66±0.01 & 0.64±0.03 & 0.63±0.00 & 0.68±0.03 & 0.66±0.00 & 0.64±0.01 & 0.65±0.00 & 0.66±0.01 & 0.65±0.00 & 0.66±0.00 & - & 0 \\
\textbf{Kidd} & 0.66±0.03 & 0.68±0.03 & 0.71±0.03 & 0.69±0.03 & 0.70±0.03 & 0.68±0.03 & 0.66±0.02 & 0.64±0.02 & 0.69±0.03 & 0.68±0.03 &  - & 0\\
\hline
\multicolumn{12}{c}{\textbf{Without labels}} \\
\hline
\textbf{Doscond} & 0.48±0.03 & \cellcolor{green!80}{0.67±0.01} & 0.65±0.01 & 0.66±0.00 & 0.55±0.02 & 0.63±0.00 & 0.61±0.01 & 0.65±0.01 & 0.61±0.01 & 0.63±0.01 & 1 & 0\\
\textbf{Kidd} & \cellcolor{green!25}{0.60±0.02} & 0.58±0.02 & 0.55±0.01 & 0.56±0.02 & 0.60±0.02 & 0.61±0.03 & 0.60±0.02 & 0.62±0.02 & 0.58±0.01 & 0.56±0.02 & 0 & 0\\
\textbf{Feature} & 0.57±0.04 & 0.60±0.05 & 0.62±0.03 & 0.65±0.03 & \cellcolor{green!80}{0.67±0.02} & 0.67±0.04 & \cellcolor{green!80}{0.70±0.02} & \cellcolor{green!25}{0.67±0.04} & \cellcolor{green!25}{0.68±0.03} & \cellcolor{green!80}{0.73±0.01} & 3 & 0 \\
\textbf{WL} & \cellcolor{green!25}{0.60±0.03} & 0.62±0.04 & \cellcolor{green!25}{0.67±0.03} & \cellcolor{green!25}{0.68±0.02} & \cellcolor{green!25}{0.65±0.03} & \cellcolor{green!25}{0.68±0.02} & \cellcolor{green!25}{0.69±0.02} & \cellcolor{green!80}{0.69±0.04} & \cellcolor{green!80}{0.69±0.02} & 0.68±0.02 & 2 & 0\\
\textbf{Random} & \cellcolor{green!80}{0.62±0.02} & 0.62±0.02 & 0.63±0.03 & 0.65±0.02 & \cellcolor{green!80}{0.67±0.02} & 0.63±0.03 & 0.67±0.02 & \cellcolor{green!25}{0.67±0.02} & \cellcolor{green!80}{0.69±0.02} & 0.68±0.02 & 3 & 0\\
\textbf{TMD} & \cellcolor{green!80}{0.62±0.02} & \cellcolor{green!25}{0.63±0.03} & \cellcolor{green!80}{0.69±0.02} & \cellcolor{green!80}{0.69±0.02} & \cellcolor{green!80}{0.67±0.02} & \cellcolor{green!80}{0.70±0.02} & 0.67±0.01 & \cellcolor{green!80}{0.69±0.02} & \cellcolor{green!80}{0.69±0.03} & \cellcolor{green!25}{0.71±0.20} & 7 & 0\\
\hline
\end{tabular}
}
\subcaption{COX2. \vspace{-.4cm}}\label{tab:COX2}
\end{subtable}

\vspace{1em}

\begin{subtable}{\textwidth}
\centering
\resizebox{1.0\textwidth}{!}{%
\begin{tabular}{l | c | c | c | c | c | c | c | c | c | c | c | c}

\hline
\multicolumn{12}{c}{\textbf{With labels}} \\
\hline
\textbf{Doscond} & 0.74±0.05 & 0.77±0.02 & 0.78±0.00 & 0.79±0.01 & 0.78±0.01 & 0.78±0.00 & 0.78±0.02 & 0.76±0.03 & 0.78±0.01 & 0.78±0.01 & - & 0 \\
\textbf{Kidd} & 0.17±0.00 & 0.17±0.00 & 0.39±0.31 & 0.83±0.00 & 0.83±0.00 & 0.83±0.00 & 0.83±0.00 & 0.83±0.00 & 0.83±0.00 & 0.39±0.31 &  - & 0 \\
\hline
\multicolumn{12}{c}{\textbf{Without labels}} \\
\hline
\textbf{Doscond} & 0.44±0.14 & 0.22±0.00 & 0.25±0.03 & 0.25±0.03 & 0.77±0.02 & 0.39±0.08 & 0.65±0.09 & 0.22±0.00 & 0.22±0.00 & 0.21±0.00 & 0 & 8\\
\textbf{Kidd} & 0.17±0.00 & \cellcolor{green!80}{0.83±0.00} & 0.17±0.00 & 0.61±0.31 & \cellcolor{green!80}{0.83±0.00} & \cellcolor{green!80}{0.83±0.00} & 0.17±0.00 & 0.17±0.00 & 0.17±0.00 & 0.61±0.31 & 3 & 5\\
\textbf{Feature} & \cellcolor{green!25}{0.65±0.06} & 0.72±0.05 & \cellcolor{green!25}{0.71±0.04} & 0.72±0.02 & \cellcolor{green!25}{0.78±0.01} & 0.75±0.03 & \cellcolor{green!25}{0.74±0.04} & 0.73±0.03 & \cellcolor{green!80}{0.78±0.02} & 0.76±0.02 & 1 & 0 \\
\textbf{WL} & 0.57±0.06 & 0.70±0.04 & \cellcolor{green!80}{0.75±0.02} & \cellcolor{green!25}{0.76±0.02} & 0.77±0.02 & \cellcolor{green!25}{0.77±0.02} & \cellcolor{green!80}{0.77±0.02} & \cellcolor{green!25}{0.77±0.02} & \cellcolor{green!80}{0.78±0.02} & \cellcolor{green!80}{0.78±0.01} & 4 & 0\\
\textbf{Random} & \cellcolor{green!25}{0.65±0.07} & 0.70±0.04 & 0.69±0.05 & 0.68±0.06 & 0.76±0.04 & 0.74±0.04 & \cellcolor{green!25}{0.74±0.03} & 0.74±0.04 & 0.75±0.06 & \cellcolor{green!25}{0.77±0.03} & 0 & 0 \\
\textbf{TMD} & \cellcolor{green!80}{0.70±0.04} & \cellcolor{green!25}{0.76±0.02} & \cellcolor{green!80}{0.75±0.03} & \cellcolor{green!80}{0.78±0.02} & \cellcolor{green!25}{0.78±0.02} & \cellcolor{green!25}{0.77±0.01} & \cellcolor{green!80}{0.77±0.01} & \cellcolor{green!80}{0.78±0.02} & \cellcolor{green!25}{0.77±0.02} & \cellcolor{green!80}{0.78±0.02} & 6 & 0 \\
\hline
\end{tabular}
}
\subcaption{PROTEINS. \vspace{-.4cm}}\label{tab:proteins}
\end{subtable}

\vspace{1em}

\begin{subtable}{\textwidth}
\centering
\resizebox{1.0\textwidth}{!}{%
\begin{tabular}{l | c | c | c | c | c | c | c | c | c | c | c| c} 
\hline
\multicolumn{12}{c}{\textbf{With labels}} \\
\hline
\textbf{Doscond} & 0.71±0.02 & 0.69±0.01 & 0.69±0.03 & 0.70±0.03 & 0.67±0.02 & 0.69±0.03 & 0.70±0.02 & 0.70±0.02 & 0.70±0.01 & 0.69±0.04 & - & 0 \\
\textbf{Kidd} & 0.68±0.00 & 0.68±0.00 & 0.68±0.00 & 0.68±0.00 & 0.68±0.00 & 0.75±0.10 & 0.68±0.00 & 0.68±0.00 & 0.68±0.00 & 0.74±0.04 & - & 0 \\
\hline
\multicolumn{12}{c}{\textbf{Without labels}} \\
\hline
\textbf{Doscond} & \cellcolor{green!25}{0.67±0.00} & \cellcolor{green!80}{0.74±0.03} & \cellcolor{green!80}{0.73±0.01} & \cellcolor{green!80}{0.73±0.01} & \cellcolor{green!25}{0.71±0.01} & \cellcolor{green!25}{0.70±0.01} & 0.70±0.00 & 0.68±0.01 & 0.72±0.01 & 0.70±0.01 & 3 & 0\\
\textbf{Kidd} & \cellcolor{green!80}{0.68±0.00} & 0.68±0.00 & 0.68±0.00 & 0.44±0.17 & 0.68±0.00 & 0.68±0.00 & 0.32±0.00 & 0.70±0.03 & 0.32±0.00 & 0.32±0.00 & 1 & 0\\
\textbf{Feature} & 0.62±0.05 & 0.65±0.06 & 0.67±0.07 & 0.69±0.03 & 0.70±0.04 & 0.68±0.04 & \cellcolor{green!25}{0.72±0.04} & \cellcolor{green!25}{0.71±0.04} & \cellcolor{green!80}{0.74±0.04} & \cellcolor{green!80}{0.74±0.04} & 2 & 3 \\
\textbf{WL} & \cellcolor{green!25}{0.67±0.04} & 0.70±0.04 & \cellcolor{green!25}{0.71±0.02} & \cellcolor{green!25}{0.70±0.03} & \cellcolor{green!25}{0.71±0.03} & \cellcolor{green!25}{0.70±0.03} & 0.68±0.03 & 0.70±0.04 & 0.70±0.04 & \cellcolor{green!80}{0.74±0.04} & 1 & 0\\
\textbf{Random} & 0.62±0.05 & 0.65±0.06 & 0.67±0.07 & 0.69±0.03 & 0.70±0.04 & 0.68±0.04 & \cellcolor{green!25}{0.72±0.04} & \cellcolor{green!25}{0.71±0.04} & \cellcolor{green!80}{0.74±0.04} & \cellcolor{green!80}{0.74±0.04} & 2 & 0\\
\textbf{TMD} & 0.64±0.07 & \cellcolor{green!25}{0.71±0.04} & 0.70±0.03 & \cellcolor{green!80}{0.73±0.04} & \cellcolor{green!80}{0.76±0.04} & \cellcolor{green!80}{0.72±0.05} & \cellcolor{green!80}{0.75±0.04} & \cellcolor{green!80}{0.75±0.03} & \cellcolor{green!25}{0.73±0.04} & \cellcolor{green!25}{0.73±0.03} & 5 & 0\\
\hline
\end{tabular}
}
\subcaption{MUTAG. \vspace{-.4cm}}\label{tab:mutag}
\end{subtable}

\vspace{1em}

\begin{subtable}{\textwidth}
\centering
\resizebox{1.0\textwidth}{!}{%
\begin{tabular}{l | c | c | c | c | c | c | c | c | c | c | c | c}
\hline
\multicolumn{12}{c}{\textbf{With labels}} \\
\hline
\textbf{Doscond} & 0.56±0.01 & 0.56±0.02 & 0.57±0.02 & 0.58±0.02 & 0.58±0.02 & 0.56±0.03 & 0.59±0.02 & 0.61±0.02 & 0.61±0.02 & 0.61±0.02 & - & 0\\
\textbf{Kidd} & 0.60±0.01 & 0.60±0.01 & 0.61±0.02 & 0.61±0.01 & 0.61±0.02 & 0.62±0.02 & 0.62±0.01 & 0.62±0.01 & 0.62±0.01 & 0.62±0.01 & - & 0 \\
\hline
\multicolumn{12}{c}{\textbf{Without labels}} \\
\hline
\textbf{Doscond} & 0.50±0.02 & 0.51±0.03 & 0.49±0.03 & 0.52±0.03 & 0.52±0.02 & 0.54±0.02 & 0.54±0.02 & 0.54±0.02 & 0.51±0.04 & 0.55±0.03 & 0 & 0\\
\textbf{Kidd} & \cellcolor{green!80}{0.55±0.03} & \cellcolor{green!25}{0.56±0.03} & \cellcolor{green!25}{0.56±0.03} & \cellcolor{green!25}{0.54±0.05} & \cellcolor{green!25}{0.57±0.02} & \cellcolor{green!25}{0.57±0.02} & \cellcolor{green!25}{0.57±0.02} & \cellcolor{green!25}{0.57±0.02} & \cellcolor{green!25}{0.58±0.02} & \cellcolor{green!25}{0.58±0.02} & 1 & 0\\
\textbf{Feature} & \cellcolor{green!25}{0.51±0.01} & 0.53±0.01 & 0.52±0.01 & 0.53±0.02 & 0.53±0.02 & 0.53±0.02 & 0.54±0.02 & 0.53±0.01 & 0.56±0.03 & 0.54±0.03 & 0 & 0\\
\textbf{WL} & \cellcolor{green!25}{0.51±0.01} & 0.51±0.01 & 0.52±0.01 & 0.50±0.01 & 0.51±0.01 & 0.51±0.01 & 0.50±0.00 & 0.51±0.01 & 0.52±0.01 & 0.53±0.02 & 0 & 0\\
\textbf{Random} & \cellcolor{green!80}{0.55±0.02} & \cellcolor{green!80}{0.59±0.02} & \cellcolor{green!80}{0.62±0.01} & \cellcolor{green!80}{0.60±0.02} & \cellcolor{green!80}{0.63±0.02} & \cellcolor{green!80}{0.62±0.02} & \cellcolor{green!80}{0.61±0.02} & \cellcolor{green!80}{0.61±0.02} & \cellcolor{green!80}{0.64±0.01} & \cellcolor{green!80}{0.65±0.01} & 10 & 0\\
\textbf{TMD} & \cellcolor{green!25}{0.51±0.01} & 0.53±0.01 & 0.52±0.01 & 0.53±0.02 & 0.53±0.02 & 0.53±0.02 & 0.54±0.02 & 0.53±0.01 & 0.56±0.03 & 0.54±0.03 & 0 & 0\\
\hline
\end{tabular}
}
\subcaption{NCI1.\vspace{-.4cm}}\label{tab:NCI1}
\end{subtable}

\end{table}

\clearpage
\newpage

\begin{table}[t!]
\centering
\caption{Graph subsampling performance across sampling percentages for OGBG datasets, reported as mean ± confidence bar of accuracy (ACC) or area under the ROC (ROC-AUC). Best and second-best per column are in dark/light green, respectively. W is total wins per method.}
\vspace{.1in}
\label{tab:all-results-graph-subsampling2}

\begin{subtable}{\textwidth}
\centering
\resizebox{1.0\textwidth}{!}{%
\begin{tabular}{l | c | c | c | c | c | c | c | c | c | c | c | c}

\hline
\multicolumn{12}{c}{\textbf{With labels}} \\
\hline
\textbf{Doscond} & 0.51±0.01 & 0.51±0.02 & 0.52±0.02 & 0.52±0.02 & 0.40±0.02 & 0.54±0.02 & 0.55±0.02 & 0.56±0.02 & 0.58±0.02 & 0.62±0.02 & - & 0 \\
\textbf{Kidd} & 0.62±0.02 & 0.62±0.04 & 0.62±0.05 & 0.62±0.06 & 0.62±0.07 & 0.58±0.15 & 0.62±0.08 & 0.63±0.10 & 0.63±0.11 & 0.63±0.12 & - & 0 \\
\hline
\multicolumn{12}{c}{\textbf{Without labels}} \\
\hline
\textbf{Doscond} & 0.43±0.02 & 0.44±0.02 & 0.30±0.02 & 0.45±0.02 & 0.46±0.03 & 0.46±0.03 & 0.32±0.01 & 0.47±0.03 & 0.47±0.04 & 0.48±0.03 & 0 & 10\\
\textbf{Kidd} & 0.57±0.04 & 0.57±0.05 & 0.58±0.04 & 0.58±0.05 & 0.58±0.06 & 0.58±0.05 & 0.58±0.06 & 0.59±0.07 & 0.59±0.07 & 0.59±0.08 & 0 & 0\\
\textbf{Feature} & \cellcolor{green!80}{0.78±0.01} & \cellcolor{green!80}{0.76±0.01} & \cellcolor{green!25}{0.76±0.04} & \cellcolor{green!80}{0.77±0.00} & \cellcolor{green!80}{0.79±0.00} & 0.77±0.01 & \cellcolor{green!25}{0.76±0.01} & \cellcolor{green!25}{0.76±0.00} & \cellcolor{green!80}{0.78±0.00} & \cellcolor{green!80}{0.78±0.01} & 6 & 0\\
\textbf{WL} & 0.57±0.08 & 0.69±0.05 & 0.73±0.06 & 0.74±0.05 & 0.76±0.01 & 0.77±0.01 & 0.72±0.06 & 0.70±0.07 & \cellcolor{green!80}{0.78±0.01} & 0.76±0.01 & 1 & 0\\
\textbf{Random} & 0.62±0.09 & \cellcolor{green!80}{0.76±0.02} & \cellcolor{green!80}{0.77±0.01} & \cellcolor{green!25}{0.76±0.01} & \cellcolor{green!25}{0.77±0.00} & \cellcolor{green!25}{0.78±0.00} & \cellcolor{green!25}{0.76±0.03} & \cellcolor{green!80}{0.77±0.00} & \cellcolor{green!25}{0.77±0.00} & \cellcolor{green!25}{0.77±0.00} & 3 & 0\\
\textbf{TMD} & \cellcolor{green!25}{0.69±0.05} & \cellcolor{green!25}{0.75±0.03} & \cellcolor{green!25}{0.76±0.02} & \cellcolor{green!25}{0.76±0.01} & 0.73±0.04 & \cellcolor{green!80}{0.79±0.02} & \cellcolor{green!80}{0.78±0.01} & \cellcolor{green!80}{0.77±0.01} & \cellcolor{green!25}{0.77±0.01} & 0.73±0.07 & 3 & 0\\
\hline
\end{tabular}
}
\subcaption{OGBG-MOLBBBP. \vspace{-.4cm}}\label{tab:molbbbp}
\end{subtable}
\vspace{1em}

\begin{subtable}{\textwidth}
\centering
\resizebox{1.0\textwidth}{!}{
\begin{tabular}{l | c | c | c | c | c | c | c | c | c | c | c | c}
\hline
\textbf{Doscond} & 0.67±0.01 & 0.67±0.03 & 0.64±0.01 & 0.67±0.01 & 0.65±0.03 & 0.68±0.02 & 0.66±0.02 & 0.67±0.01 & 0.67±0.04 & 0.66±0.02 & - & 0 \\
\textbf{Kidd} & 0.64±0.03 & 0.66±0.03 & 0.65±0.03 & 0.64±0.03 & 0.61±0.03 & 0.63±0.03 & 0.66±0.03 & 0.67±0.03 & 0.69±0.03 & 0.68±0.03 & - & 0 \\
\hline
\multicolumn{12}{c}{\textbf{Without labels}} \\
\hline
\textbf{Doscond} & \cellcolor{green!80}{0.53±0.02} & 0.45±0.04 & \cellcolor{green!80}{0.64±0.03} & \cellcolor{green!80}{0.60±0.02} & 0.37±0.01 & \cellcolor{green!80}{0.56±0.06} & \cellcolor{green!80}{0.61±0.02} & 0.54±0.01 & \cellcolor{green!80}{0.59±0.03} & \cellcolor{green!80}{0.62±0.03} & 7 & 2\\
\textbf{Kidd} & \cellcolor{green!80}{0.53±0.03} & \cellcolor{green!25}{0.51±0.03} & 0.46±0.02 & 0.48±0.02 & \cellcolor{green!80}{0.56±0.03} & \cellcolor{green!25}{0.54±0.03} & \cellcolor{green!25}{0.58±0.03} & 0.55±0.02 & 0.50±0.03 & 0.52±0.03 & 2 & 2\\
\textbf{Feature} & \cellcolor{green!25}{0.50±0.03} & \cellcolor{green!80}{0.52±0.03} & 0.53±0.02 & \cellcolor{green!25}{0.55±0.02} & \cellcolor{green!25}{0.55±0.02} & 0.51±0.02 & 0.55±0.02 & \cellcolor{green!25}{0.56±0.03} & 0.56±0.02 & 0.56±0.02 & 1 & 0\\
\textbf{WL} & 0.48±0.01 & 0.49±0.02 & 0.48±0.02 & 0.51±0.03 & 0.48±0.02 & 0.50±0.02 & 0.53±0.03 & 0.50±0.03 & 0.55±0.03 & 0.53±0.03 & 0 & 4 \\
\textbf{Random} & 0.49±0.02 & 0.49±0.03 & 0.51±0.03 & 0.49±0.02 & 0.48±0.03 & 0.50±0.04 & 0.48±0.02 & 0.48±0.02 & 0.49±0.03 & 0.51±0.03 & 0 & 7\\
\textbf{TMD} & 0.45±0.01 & \cellcolor{green!80}{0.52±0.03} & \cellcolor{green!25}{0.55±0.04} & 0.49±0.02 & 0.54±0.03 & \cellcolor{green!25}{0.54±0.03} & \cellcolor{green!25}{0.58±0.03} & \cellcolor{green!80}{0.60±0.03} & \cellcolor{green!25}{0.57±0.02} & \cellcolor{green!25}{0.58±0.02} & 2 & 2\\
\hline
\end{tabular}
}
\subcaption{OGBG-MOLBACE. \vspace{-.4cm}}\label{tab:molbace}
\end{subtable}
\vspace{1em}

\end{table}

\begin{table}[H]
\scriptsize
\caption{Node subsampling performance across datasets and sampling percentages, reported as mean ± confidence bar of accuracy (ACC). Best and second-best per column are in dark and light green, respectively. W is total wins per method.}
\label{tab:combined-results-nodes}
\vspace{0.1in}
\centering

\begin{subtable}{\textwidth}
\centering
\resizebox{1.0\textwidth}{!}{%
\begin{tabular}{l|c|c|c|c|c|c|c|c|c|c}
\hline
\textbf{Method} & \textbf{1\%} & \textbf{2\%} & \textbf{3\%} & \textbf{4\%} & \textbf{5\%} & \textbf{6\%} & \textbf{7\%} & \textbf{8\%} & \textbf{9\%} & \textbf{W} \\
\hline
\textbf{RW}
    & \cellcolor{green!80}{0.68±0.02}
    & \cellcolor{green!80}{0.67±0.03}
    & 0.66±0.03
    & 0.65±0.04
    & 0.66±0.04
    & 0.68±0.08
    & 0.71±0.05
    & \cellcolor{green!25}{0.77±0.03}
    & 0.80±0.04
    & 2 \\
\textbf{k-cores}
    & 0.56±0.07
    & \cellcolor{green!25}{0.63±0.04}
    & 0.48±0.07
    & 0.64±0.06
    & \cellcolor{green!25}{0.67±0.05}
    & 0.65±0.06
    & \cellcolor{green!25}{0.73±0.04}
    & 0.73±0.03
    & 0.74±0.03
    & 0 \\
\textbf{Random}
    & \cellcolor{green!80}{0.68±0.04}
    & \cellcolor{green!80}{0.67±0.03}
    & \cellcolor{green!80}{0.70±0.03}
    & \cellcolor{green!80}{0.71±0.03}
    & 0.66±0.03
    & \cellcolor{green!25}{0.69±0.03}
    & \cellcolor{green!80}{0.74±0.03}
    & 0.75±0.03
    & \cellcolor{green!25}{0.81±0.04}
    & 4 \\
\textbf{TMD}
    & \cellcolor{green!25}{0.66±0.05}
    & \cellcolor{green!80}{0.67±0.03}
    & \cellcolor{green!25}{0.68±0.04}
    & \cellcolor{green!25}{0.69±0.02}
    & \cellcolor{green!80}{0.68±0.04}
    & \cellcolor{green!80}{0.72±0.03}
    & \cellcolor{green!80}{0.74±0.04}
    & \cellcolor{green!80}{0.78±0.03}
    & \cellcolor{green!80}{0.84±0.02}
    & 6 \\
\hline
\end{tabular}
}
\caption{MUTAG}
\label{tab:mutag-pivot}
\end{subtable}

\vspace{0em}

\begin{subtable}{\textwidth}
\centering
\resizebox{1.0\textwidth}{!}{%
\begin{tabular}{l|c|c|c|c|c|c|c|c|c|c}

\hline
\textbf{RW}
    & \cellcolor{green!80}{0.61±0.01}
    & \cellcolor{green!80}{0.60±0.01}
    & \cellcolor{green!80}{0.63±0.01}
    & 0.63±0.01
    & \cellcolor{green!25}{0.67±0.02}
    & \cellcolor{green!25}{0.69±0.01}
    & \cellcolor{green!25}{0.68±0.02}
    & \cellcolor{green!80}{0.72±0.01}
    & 0.70±0.02
    & 4 \\
\textbf{k-cores}
    & 0.59±0.01
    & \cellcolor{green!80}{0.60±0.01}
    & 0.60±0.01
    & \cellcolor{green!25}{0.64±0.02}
    & 0.65±0.01
    & \cellcolor{green!25}{0.69±0.02}
    & \cellcolor{green!25}{0.68±0.02}
    & 0.70±0.01
    & \cellcolor{green!25}{0.71±0.01}
    & 1 \\
\textbf{Random}
    & \cellcolor{green!25}{0.60±0.01}
    & \cellcolor{green!80}{0.60±0.01}
    & \cellcolor{green!25}{0.62±0.02}
    & \cellcolor{green!25}{0.64±0.02}
    & \cellcolor{green!25}{0.67±0.02}
    & 0.68±0.01
    & \cellcolor{green!80}{0.71±0.01}
    & 0.70±0.02
    & 0.70±0.02
    & 2 \\
\textbf{TMD}
    & \cellcolor{green!25}{0.60±0.01}
    & \cellcolor{green!80}{0.60±0.01}
    & \cellcolor{green!80}{0.63±0.01}
    & \cellcolor{green!80}{0.65±0.01}
    & \cellcolor{green!80}{0.68±0.01}
    & \cellcolor{green!80}{0.70±0.01}
    & \cellcolor{green!80}{0.71±0.01}
    & \cellcolor{green!25}{0.71±0.01}
    & \cellcolor{green!80}{0.73±0.01}
    & 7 \\
\hline
\end{tabular}
}
\caption{PROTEINS}
\label{tab:proteins-pivot}
\end{subtable}

\vspace{0em}

\begin{subtable}{\textwidth}
\centering
\resizebox{1.0\textwidth}{!}{%
\begin{tabular}{l|c|c|c|c|c|c|c|c|c|c}

\hline
\textbf{RW}
    & \cellcolor{green!25}{0.58±0.01}
    & \cellcolor{green!80}{0.59±0.01}
    & \cellcolor{green!80}{0.61±0.01}
    & 0.63±0.01
    & \cellcolor{green!80}{0.66±0.02}
    & \cellcolor{green!80}{0.69±0.01}
    & \cellcolor{green!80}{0.72±0.02}
    & \cellcolor{green!25}{0.73±0.02}
    & \cellcolor{green!25}{0.74±0.02}
    & 5 \\
\textbf{k-cores}
    & \cellcolor{green!80}{0.59±0.01}
    & \cellcolor{green!80}{0.59±0.01}
    & \cellcolor{green!25}{0.59±0.02}
    & 0.59±0.01
    & 0.60±0.01
    & \cellcolor{green!25}{0.59±0.01}
    & 0.59±0.01
    & 0.60±0.01
    & 0.59±0.01
    & 2 \\
\textbf{Random}
    & \cellcolor{green!80}{0.59±0.01}
    & \cellcolor{green!25}{0.58±0.01}
    & \cellcolor{green!80}{0.61±0.01}
    & \cellcolor{green!25}{0.64±0.01}
    & \cellcolor{green!80}{0.66±0.02}
    & \cellcolor{green!80}{0.69±0.02}
    & \cellcolor{green!25}{0.71±0.02}
    & \cellcolor{green!25}{0.73±0.02}
    & 0.73±0.03
    & 7 \\
\textbf{TMD}
    & \cellcolor{green!25}{0.58±0.02}
    & \cellcolor{green!80}{0.59±0.01}
    & \cellcolor{green!80}{0.61±0.01}
    & \cellcolor{green!80}{0.65±0.02}
    & \cellcolor{green!25}{0.65±0.02}
    & \cellcolor{green!80}{0.69±0.02}
    & 0.70±0.02
    & \cellcolor{green!80}{0.74±0.01}
    & \cellcolor{green!80}{0.76±0.01}
    & 6 \\
\hline
\end{tabular}
}
\caption{DD}
\label{tab:dd-pivot}
\end{subtable}

\vspace{0em}

\begin{subtable}{\textwidth}
\centering
\resizebox{1.0\textwidth}{!}{%
\begin{tabular}{l|c|c|c|c|c|c|c|c|c|c}

\hline
\textbf{RW}
    & \cellcolor{green!25}{0.78±0.01}
    & 0.77±0.01
    & \cellcolor{green!80}{0.79±0.01}
    & \cellcolor{green!80}{0.80±0.02}
    & \cellcolor{green!80}{0.79±0.02}
    & \cellcolor{green!80}{0.78±0.02}
    & \cellcolor{green!25}{0.79±0.02}
    & \cellcolor{green!25}{0.79±0.02}
    & 0.77±0.02
    & 4 \\
\textbf{k-cores}
    & \cellcolor{green!25}{0.78±0.01}
    & \cellcolor{green!25}{0.78±0.02}
    & 0.77±0.01
    & \cellcolor{green!25}{0.79±0.02}
    & 0.74±0.05
    & \cellcolor{green!80}{0.78±0.01}
    & 0.77±0.02
    & 0.78±0.02
    & \cellcolor{green!80}{0.79±0.02}
    & 2 \\
\textbf{Random}
    & \cellcolor{green!80}{0.79±0.02}
    & \cellcolor{green!25}{0.78±0.02}
    & \cellcolor{green!25}{0.78±0.01}
    & \cellcolor{green!25}{0.79±0.02}
    & \cellcolor{green!80}{0.79±0.01}
    & \cellcolor{green!80}{0.78±0.02}
    & 0.77±0.02
    & 0.78±0.01
    & \cellcolor{green!25}{0.78±0.01}
    & 3 \\
\textbf{TMD}
    & \cellcolor{green!80}{0.79±0.02}
    & \cellcolor{green!80}{0.79±0.02}
    & \cellcolor{green!25}{0.78±0.01}
    & 0.77±0.02
    & \cellcolor{green!25}{0.78±0.01}
    & \cellcolor{green!80}{0.78±0.02}
    & \cellcolor{green!80}{0.80±0.02}
    & \cellcolor{green!80}{0.80±0.01}
    & \cellcolor{green!80}{0.79±0.02}
    & 6 \\
\hline
\end{tabular}
}
\caption{COX2}
\label{tab:cox2-pivot}
\end{subtable}

\end{table}

%\FloatBarrier

\section{Discussion}

We proposed the first theoretically grounded approaches for subsampling graph datasets that preserve the performance of a downstream GNN. Our methods apply to both graph and node subsampling. By leveraging the TMD, our methods ensure that the subsampled data maintain structural similarities with the original dataset. Thus, we can bound the loss incurred by subsampling. Our experiments demonstrate our method outperforms other methods for these problems. 

Future work could combine node and graph subsampling to optimize tradeoffs between storage, computational efficiency, and accuracy. While our analysis is tailored to graph-level tasks, extending these techniques to node-level tasks, where graphs are often much larger, is a promising direction. Additionally, exploring TMD's implications for edge-level tasks, such as link prediction, could further enhance its theoretical and practical utility.

\section*{Acknowledgements}
The authors thank Ching-Yao Chuang and Joshua Robinson for valuable discussions.  This work was supported in part by NSF grant CCF-2338226, the NSF AI Institute TILOS and an Alexander von Humboldt fellowship.

%\FloatBarrier

\bibliography{ref}
\bibliographystyle{plainnat}

\newpage
\appendix

\section{Additional background}\label{sec:additional-background}

In this Appendix, we discuss additional helpful background for the discussions in the main body. 

\subsection{Additional details about the TMD}\label{app:TMD}

Here we describe the padding function $\rho$ using in $\cOTbar$.

We use $\blankTree^n$ to denote $n$ disjoint copies of a blank tree $\blankTree.$ Given two tree multisets $\cT_u(T)$, $\cT_v(T')$, we define $\rho$ to be the following augmentation function, which returns two multi-sets of the same size: 
\begin{align}\label{eq:rho}
    \rho: (\cT_v(T'), \cT_u(T)) \mapsto \paren{\cT_v(T') \cup {\blankTree}^{\max(|\cT_u(T)| - |\cT_v(T')|, 0)}, \cT_u(T) \cup {\blankTree}^{\max(|\cT_v(T)| - |\cT_u(T')|, 0)} }.
\end{align}

Equipped with this definition, we define $\cOTbar$ for a given weight function $w: \N \to \R_{> 0}$ as follows: 
\begin{align}\label{eq:OT-bar}
    \cOTbar(\cdot, \cdot) = \cOT_{\cTD_w}(\rho(\cdot, \cdot)). 
\end{align}

\subsection{Visual notation aids}\label{sec:notation}

\begin{figure}
    \centering
    \includegraphics[width=0.75\linewidth]{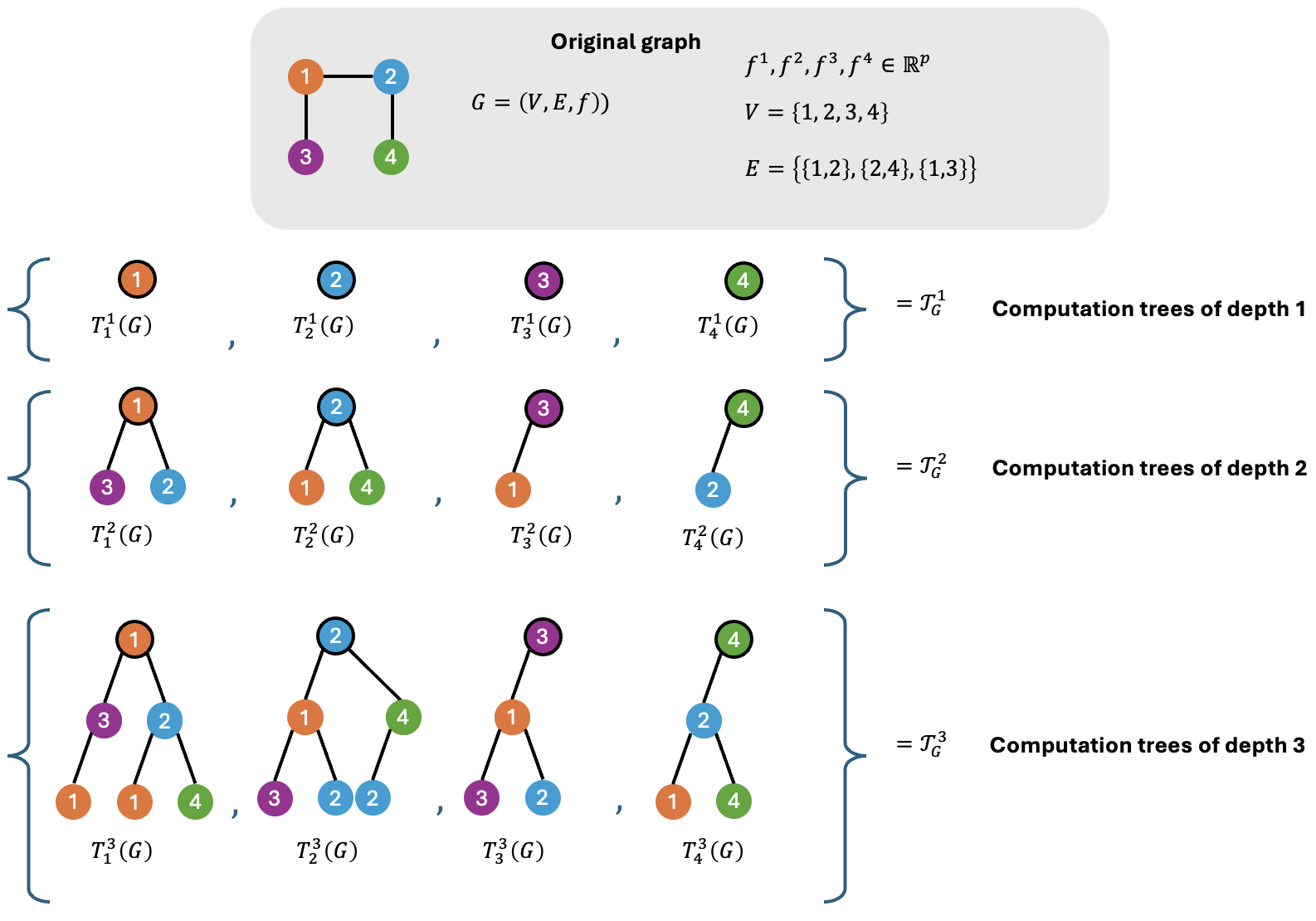}
    \caption{Computation trees up to depth $L=3$ for an example 4-node graph (Definition~\ref{def:tree}).}
    \label{fig:ctrees}
\end{figure}

\begin{figure}
    \centering
    \includegraphics[width=0.4\linewidth]{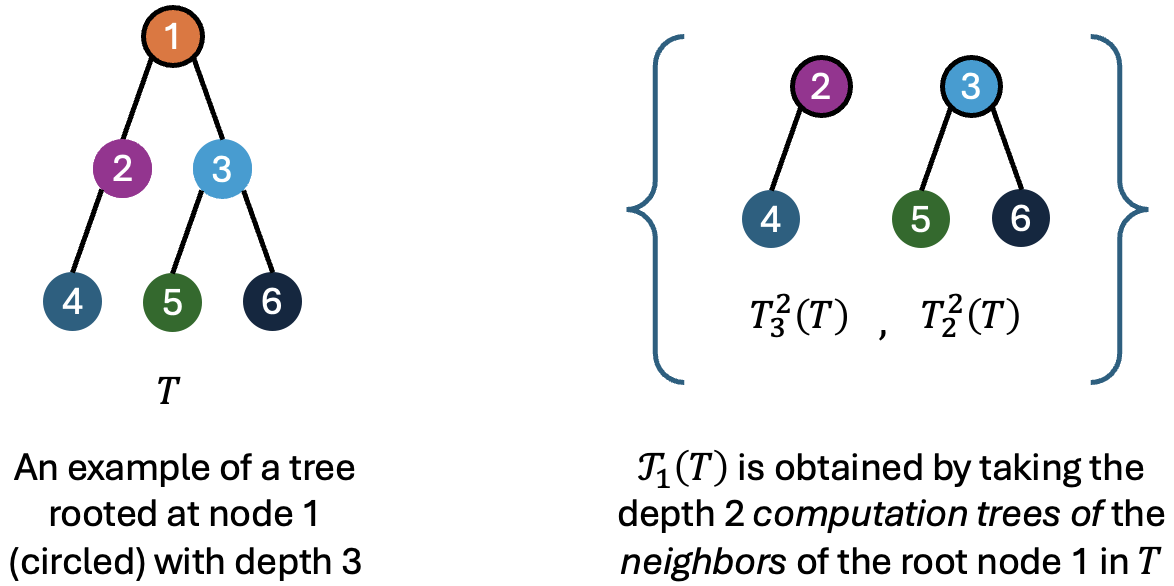}
    \caption{The tree multiset associated with an example depth 3 rooted tree (Definition~\ref{def:multiset}).}
    \label{fig:multiset}
\end{figure}

\begin{figure}[ht]
    \centering
    \includegraphics[width=1\linewidth]{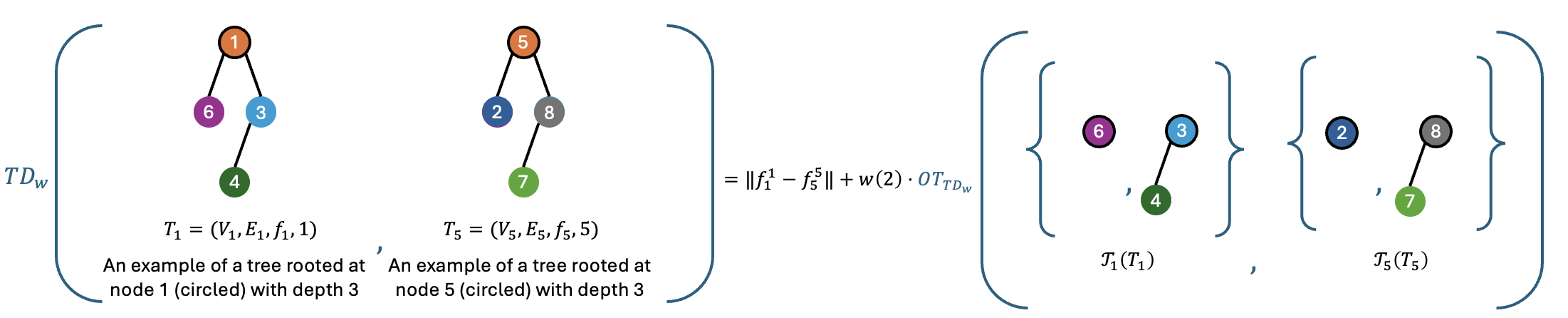}
    \caption{The tree distance between example graphs (Definition~\ref{def:TD}).}
    \label{fig:td}
\end{figure}

\begin{figure}[ht]
    \centering
    \includegraphics[width=.25\linewidth]{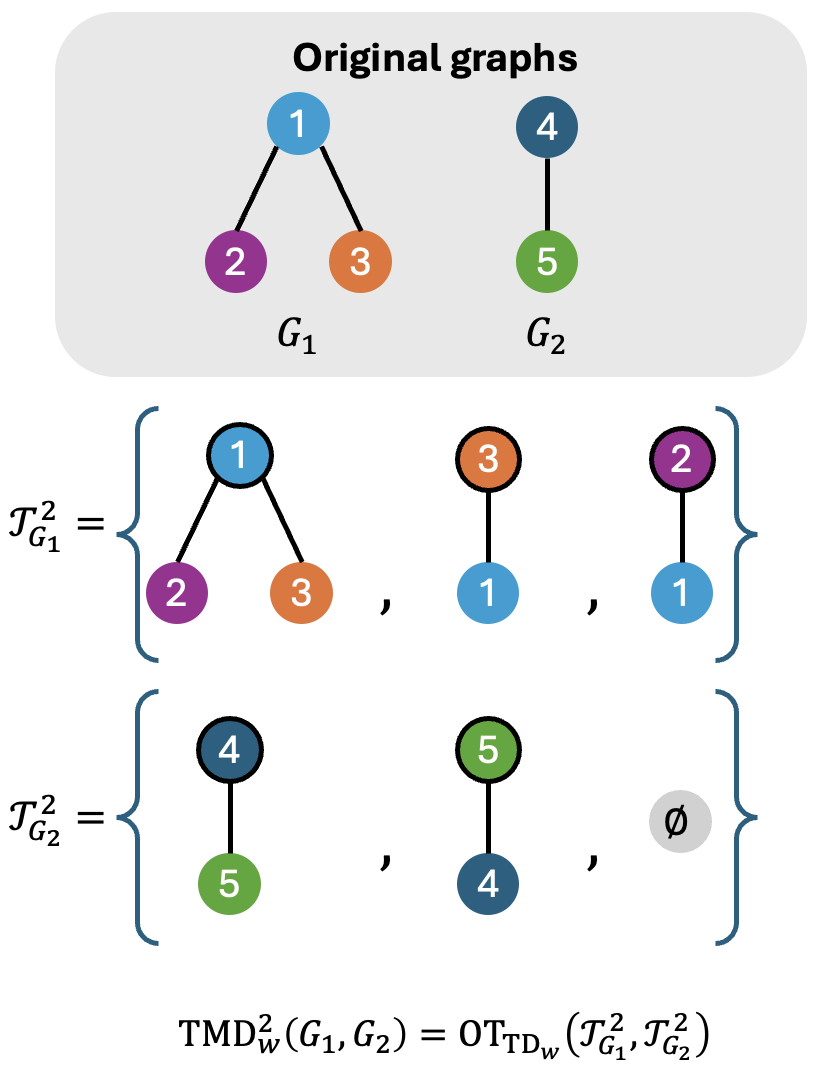}
    \caption{TMD computation between example graphs (Definition~\ref{def:TMD}). We first construct the computation trees of the original graphs, followed by computing the OT cost with respect to the tree distance. Since the graphs have unequal number of nodes, we pad the computation trees of $G_2$ with a \emph{blank} tree (see also \eqref{eq:rho}).}
    \label{fig:tmd}
\end{figure}

We have included some visualizations to aid in understanding the notations introduced in Section~\ref{sec:prelim}. In the following figures, we use black outlines to denote the roots of rooted trees.  Figure~\ref{fig:ctrees} gives a visualization of computation trees for a simple four-node graph, as per Definition~\ref{def:tree}. Figure~\ref{fig:multiset} gives a visualization of a tree multiset as per Definition~\ref{def:multiset}. Figure~\ref{fig:td} gives a visualization of the recursive definition of the tree distance (TD) Definition~\ref{def:TD}. Figure~\ref{fig:tmd} shows a visualization of computing TMD as an OT problem between multiset of computation trees of two graphs. We hope that these figures help the reader to develop a more intuitive understanding of the TMD.

\subsection{Heuristic for selecting $\cS$ in the relaxed node subsampling problem}\label{sec:heuristic}

Here we present
our heuristic for constructing the set of candidate subsets $\cS$ in our experiments. Most real-world networks are scale-free and small-world, and hence can be well-modeled by random graphs, such as preferential attachment, configuration models, and {inhomogeneous} random graphs \citep{bollobas2004coupling,newman1999scaling}. {These} 
graphs
converge \textit{locally} to {graph limits} \citep[Vol.~2, Ch.~2]{van2024random}. By a ``transitive'' argument, {we can} view real-world networks as parts of sequences converging to graph limits in the same sense. 
\citet{alimohammadi2023local} showed that for 
such graphs, sampling {nodes'} {local} breadth-first search (BFS) trees asymptotically preserves critical motifs of the {graph limit} (see Appendix~\ref{sec:random-graph-limits} for an overview). Thus, in our experiments, to construct our candidate subset $\cS$ for the \emph{relaxed node subsampling problem}, we include the BFS search trees rooted at the graph's nodes up to a certain node budget.  

\begin{definition}[$k$-BFS subset] {Given $G = (V, E, f)$ and $v \in V$}, let $\ell_k$ be the deepest level such that the $v$-rooted BFS tree of depth $\ell_k$ has at most $k$ nodes (breaking ties in a fixed but arbitrary way). {The $k$-BFS subset of $v$, denoted $S_{\mathrm{BFS}(v;k)}$, is} the set of nodes at distance $\leq \ell_{k}$ from $v$ in $G$.
\end{definition}

In addition, we augment the candidate subsets with additional candidate subgraphs proposed in prior random-walk-inspired heuristics based on random walks (RW) \citep{razin2023ability} and graph cores (k-cores) \citep{salha2022degeneracy}. Concretely, given a node budget $k$, \citet{razin2023ability} provide an algorithm to construct a subset $S_{\mathrm{RW}}$ of size at most $k$; and \citet{salha2022degeneracy} provide an algorithm to construct a subset $S_{\mathrm{k-core}}$ of size at most $k$. 

Combining these three heuristics, we use: 
\begin{align*}
    \cS = \cup_{v \in V} S_{\mathrm{BFS}(v;k)} \cup \{S_{\mathrm{RW}}, S_{\mathrm{k-core}}\} 
\end{align*}
as our candidate subsets for the \emph{relaxed node subsampling problem} in our node subsampling experiments. We then apply Theorem~\ref{thm:node-subsampling-relaxed} to select among these candidate node subsets in $\cS$ using the TMD. Note that $\cS$ contains a total of $|V| + 2 = O(|V|)$ graphs, each consisting of $k$ nodes. Thus, using Theorem~\ref{thm:node-subsampling-relaxed} and Theorem~\ref{thm:node-subsampling-relaxed-algorithm}, we obtain an overall runtime of $O(L |V| |E|)$ for our node subsampling procedure.

\subsection{Random graph limits}\label{sec:random-graph-limits}

As we discussed in Section~\ref{sec:heuristic}, most real networks are scale-free and small-world, and hence well-modeled by random graph models, such as preferential attachment, configuration models and inhomogenous random graphs. As shown by \citet{van2024random}, such random graph models produce graphs that can be shown to converge \textit{locally} to a limit $(G,o)$, where $G$ is a graph to which we assign a root node $o$. By a ``transitive'' argument, it makes sense to see real networks as parts of sequences converging to graph limits in a similar sense. 

To be precise, let $\mathcal{G}_*$ be the set of all possible rooted graphs. A limit graph is defined as a measure over the space $\mathcal{G}_*$ with respect to the local metric
\begin{align*}
    d_{loc}((G_1,o_1),(G_2,o_2)) = \frac{1}{1+\inf_k\{k:B_k(G_1,o_1)\not\simeq B_k(G_2,o_2)\}}
\end{align*}
 where $B_k(G,v)$ is the $k$-hop neighborhood of node $v$, and $\simeq$ is the graph isomorphism. A sequence of graphs converging to this limit is defined as follows.

\begin{definition}[Local convergence \citep{alimohammadi2023local}] Let $G_n = (V_n, E_n)$ denote a finite connected graph. Let $(G_n, o_n)$ be the rooted graph obtained by letting $o_n \in V_n$ be chosen uniformly at random. We say that $(G_n, o_n)$ converges locally to the connected rooted graph $(G, o)$, which is a (possibly random) element of $\mathcal{G}_*$ having law $\mu$, when, for every bounded and continuous function $h: \mathcal{G}_* \to \mathbb{R}$,
$$
\mathbb{E}[h(G_n,o_n)] \to \mathbb{E}_\mu(G,o)
$$
where the expectation on the right-hand-side is with respect to $(G, o)$ having law $\mu$, while the expectation on the left-hand-side is with respect to the random vertex $o_n$.
\end{definition}

\begin{table}[ht]
\centering
\caption{Overview of Graph Datasets}
\vskip 0.1in
\begin{tabular}{lccccc}
\toprule
\textbf{Dataset} & \textbf{\#Graphs} & \textbf{Avg. Nodes} & \textbf{Avg. Edges} & \textbf{\#Classes} & \textbf{Domain}\\
\midrule
MUTAG & 188 & 17.93 & 19.79 & 2 & Molecules\\
PROTEINS & 1,113 & 39.06 & 72.82 & 2 & Molecules \\
COX2 & 467 & 43.45 & 43.45 & 2 & Molecules \\
NCI1 & 4,110 & 29.87 & 32.30 & 2 & Molecules \\
DD & 1,178 & 284.32 & 715.66 & 2 & Proteins \\
OGBG-MOLBACE & 1,513 & 34.1 & 36.9 & 2 & Molecules \\
OGBG-MOLBBBP & 2,039 & 24.1 & 26.0 & 2 & Molecules \\
\bottomrule
\end{tabular}
\label{table:graph_datasets_summary}
\end{table}

\section{Additional experiments and experimental details}\label{sec:additional-background-experiments}

In this Appendix, we cover additional experimental details and experimental results to complement our discussion in the main body. 

\subsection{Details about datasets considered in empirical evaluation}\label{sec:dataset_details}

In Table~\ref{table:graph_datasets_summary}, we provide statistics pertaining to the datasets we consider in our empirical study. 

\subsection{Additional details regarding experimental setup for Table~\ref{tab:all-results-graph-subsampling1} and \ref{tab:all-results-graph-subsampling2}}\label{sec:additional-experiment-details}

All experiments were run on an NIVIDIA A6000 GPU with 1TB of RAM. The GNNs were implemented using Pytorch Geometric \citep{he2024pytorch}.  The TMD weight function was set according to Pascal's triangle rule~\citep[][Theorem 8]{Chuang22:Tree} and our implementation of TMD was based on \citet{Chuang22:Tree}. We used the Graph Kernel Library \citep{siglidis2020grakel} for the kernel distances in our experiments.  We will make the code for all of our experiments publicly available if the paper is accepted. For now, we have included an anonymous repository link in the main body of the paper. 

For the experiments in Table~\ref{tab:all-results-graph-subsampling1} using TUDatasets \citep{morris2020tudataset} (MUTAG, PROTEINS, COX2, NCI1), we trained a graph isomorphism network (GIN) with three layers. The model was optimized using the Adam optimizer with a learning rate of 0.01 and a binary cross-entropy (BCE) loss with logits. The batch size was set to 16 for TUD datasets and 64 for OGBG datasets. Each layer contained 128 hidden channels for TUD datasets and 256 for OGBG datasets. We used a global add pooling function for aggregation, with no weight regularization or dropout applied. Additionally, batch normalization was not used in the model.

For the TUDatasets, this is the same architecture used in the original paper by \citet{Chuang22:Tree} for evaluating the TMD as a measure of GNN robustness, since all of their experiments were also on TUDatasets. For the OGBG datasets, (OGBG-MOLBACE, OGBG-MOLBBBP) because they are significantly larger in terms of the \emph{number} of graphs, we use the same architecture and hyperparameters with the exception that we set the batch size to 64 to accommodate the large number of graphs and increase training efficiency, and we used a larger number of hidden channels (2x) to handle the larger, more complex distribution over graphs. We refrained from using batch normalization in both models to best align with our theoretical analysis. Models were implemented using standard GIN implementations in PytorchGeometric \citep{fey2019fast}. We focus on the GIN architecture because it is known to come with strong theoretical expressiveness guarantees and is a common choice for achieving state-of-the-art message-passing GNN performance \citep{xu19gin}. However, as the results of \citet{Chuang22:Tree} can be extended to other architectures, such as Graph Convolutional Networks \citep{kipf2016semi}, our results can easily be extended to other message-passing architectures. 

\subsection{Definition of feature-medoids distance metric}\label{sec:additional-background-experiments-feature}

For the ``Feature'' rows of our empirical results we use the following feature-distance function for graphs $G_1 = (V_1, E_1, f_1), G_2 = (V_2, E_2, f_2)$
\begin{align*}
    D_{\mathrm{feature}}(G_1, G_2) \defeq \norm{\frac{1}{\abs{V_1}} \sum_{v \in V_1} f_1 - \frac{1}{\abs{V_2}}\sum_{v \in V_2} f_2}_2. 
\end{align*}
To interpret this formula, note that it is essentially the Euclidean distance between the \emph{average} feature value in $G_1$ and the \emph{average} feature value in $G_2$ where the average is taken across all nodes in the graph. We then run $k$-medoids clustering in the distance metric $D_\mathrm{feature}$ to construct our subsampled graph datasets.

\subsection{Effects of validation and label usage in KiDD and DosCond}\label{sec:additional-background-experiments-vallabel}

The original KiDD and DosCond methods rely on access to graph labels for both the training dataset and a held-out validation dataset. In contrast, our approach is fully label-agnostic, requiring neither a labeled validation dataset nor labeled training data. To ensure a fair comparison, we adapt KiDD and DosCond by removing their dependence on labeled validation and training datasets. Specifically, we eliminate the validation set by selecting the GNN at the final epoch, after the default number of epochs used for the original model, rather than relying on validation-based model selection. We also remove the need for training labels by modifying the loss function to exclude label-dependent terms and initializing the sampled graphs randomly instead of using class-dependent initialization.

These adapted versions of KiDD and DosCond, which operate without labeled validation and training data, serve as the most direct comparison to our label-agnostic approach. As shown in Tables~\ref{tab:kidd} and~\ref{tab:doscond}, the removal of validation and label information in these methods each results in a measurable decline in performance. The simultaneous removal of both further exacerbates this decline, highlighting the challenges of achieving strong performance in fully label-agnostic settings. These results underscore the performance of our approach, which achieves competitive outcomes under label constraints.

\begin{table}[ht]
\centering
\caption{Performance comparison of KiDD across multiple datasets and percentages of graphs sampled. Performance is reported as mean ± standard deviation of accuracy (ACC) and area under the receiver operating curve (ROC-AUC), for configurations of KiDD with and without labels (lab.) and validation (val.).}
\vskip .1in
\label{tab:kidd}
\resizebox{1.0\textwidth}{!}{%
\begin{tabular}{cc cccccccccc}
\toprule
\multicolumn{2}{c}{} & \multicolumn{10}{c}{\textbf{\% of graphs sampled}} \\
\cmidrule(lr){3-12}
\textbf{lab.} & \textbf{val.} 
& \textbf{1} & \textbf{2} & \textbf{3} & \textbf{4} & \textbf{5} 
& \textbf{6} & \textbf{7} & \textbf{8} & \textbf{9} & \textbf{10} \\
\midrule
\multicolumn{12}{c}{\textbf{MUTAG} (ACC)} \\
\midrule

\xmark & \xmark
& 0.684±0.000 & 0.684±0.000 & 0.684±0.000 & 0.439±0.174 & 0.684±0.000
& 0.684±0.000 & 0.316±0.000 & 0.702±0.025 & 0.316±0.000 & 0.316±0.000 \\

\checkmark & \xmark
& 0.684±0.000 & 0.316±0.000 & 0.684±0.020 & 0.684±0.000 & 0.684±0.000
& 0.684±0.000 & 0.316±0.000 & 0.700±0.030 & 0.316±0.000 & 0.684±0.010 \\

\xmark & \checkmark
& 0.684±0.000 & 0.684±0.000 & 0.684±0.000 & 0.684±0.000 & 0.684±0.000
& 0.684±0.000 & 0.684±0.000 & 0.684±0.000 & 0.684±0.000 & 0.684±0.000 \\

\checkmark & \checkmark
& 0.684±0.000 & 0.684±0.000 & 0.684±0.000 & 0.684±0.000 & 0.684±0.000
& 0.754±0.099 & 0.684±0.000 & 0.684±0.000 & 0.684±0.000 & 0.737±0.043 \\
\midrule

\multicolumn{12}{c}{\textbf{COX2} (ACC)} \\
\midrule

\xmark & \xmark
& 0.170±0.000 & 0.170±0.000 & 0.390±0.311 & 0.830±0.000 & 0.830±0.000
& 0.830±0.000 & 0.830±0.000 & 0.830±0.000 & 0.830±0.000 & 0.390±0.311 \\

\checkmark & \xmark
& 0.610±0.311 & 0.390±0.311 & 0.500±0.200 & 0.610±0.311 & 0.830±0.000
& 0.830±0.000 & 0.610±0.311 & 0.390±0.311 & 0.390±0.311 & 0.610±0.311 \\

\xmark & \checkmark
& 0.390±0.311 & 0.170±0.170 & 0.170±0.000 & 0.170±0.000 & 0.830±0.000
& 0.170±0.000 & 0.170±0.000 & 0.830±0.000 & 0.170±0.000 & 0.830±0.000 \\

\checkmark & \checkmark
& 0.170±0.000 & 0.830±0.000 & 0.170±0.000 & 0.610±0.311 & 0.830±0.000
& 0.830±0.000 & 0.170±0.000 & 0.170±0.000 & 0.170±0.000 & 0.610±0.311 \\
\midrule

\multicolumn{12}{c}{\textbf{NCI1} (ACC)} \\
\midrule

\xmark & \xmark
& 0.550±0.030 & 0.555±0.028 & 0.560±0.026 
& 0.540±0.050  & 0.565±0.024 & 0.568±0.022 & 0.570±0.023 & 0.574±0.022 & 0.578±0.020 & 0.580±0.021 \\

\checkmark & \xmark
& 0.572±0.020 & 0.578±0.022 & 0.582±0.023 & 0.585±0.020 & 0.588±0.021
& 0.590±0.019 & 0.591±0.020 & 0.593±0.022 & 0.595±0.018 & 0.598±0.021 \\

\xmark & \checkmark
& 0.585±0.020 
& 0.552±0.044 
& 0.590±0.018 & 0.592±0.013 & 0.593±0.017
& 0.595±0.015 & 0.596±0.012 & 0.598±0.016 & 0.599±0.014 & 0.600±0.013 \\

\checkmark & \checkmark
& 0.602±0.013 & 0.605±0.010 & 0.608±0.015 & 0.612±0.012 & 0.613±0.016
& 0.615±0.015 & 0.615±0.013 & 0.617±0.010 & 0.619±0.011 & 0.621±0.012 \\
\midrule

\multicolumn{12}{c}{\textbf{PROTEINS} (ACC)} \\
\midrule

\xmark & \xmark
& 0.600±0.023 & 0.580±0.018 & 0.550±0.013 & 0.560±0.018 & 0.600±0.020
& 0.610±0.028 & 0.600±0.023 & 0.620±0.023 & 0.580±0.013 & 0.560±0.020 \\

\checkmark & \xmark
& 0.560±0.020 & 0.610±0.030 & 0.590±0.030 & 0.590±0.025 & 0.600±0.025
& 0.620±0.030 & 0.620±0.025 & 0.600±0.025 & 0.610±0.030 & 0.610±0.025 \\

\xmark & \checkmark
& 0.640±0.020 & 0.660±0.030 & 0.660±0.025 & 0.650±0.020 & 0.640±0.030
& 0.620±0.025 & 0.630±0.030 & 0.660±0.020 & 0.680±0.030 & 0.670±0.025 \\

\checkmark & \checkmark
& 0.660±0.030 & 0.680±0.025 & 0.710±0.030 & 0.690±0.030 & 0.700±0.030
& 0.680±0.025 & 0.660±0.020 & 0.640±0.020 & 0.690±0.030 & 0.680±0.025 \\
\midrule

\multicolumn{12}{c}{\textbf{OGBG-molbace} (ROC-AUC)} \\
\midrule

\xmark & \xmark
& 0.53±0.03 & 0.51±0.03 & 0.46±0.02 & 0.48±0.02 & 0.56±0.03
& 0.54±0.03 & 0.58±0.03 & 0.55±0.02 & 0.50±0.03 & 0.52±0.03 \\

\checkmark & \xmark
& 0.58±0.03 & 0.56±0.03 & 0.61±0.02 & 0.62±0.03 & 0.60±0.03
& 0.59±0.03 & 0.61±0.02 & 0.62±0.03 & 0.58±0.03 & 0.56±0.03 \\

\xmark & \checkmark
& 0.60±0.03 & 0.59±0.03 & 0.56±0.03 & 0.58±0.03 & 0.61±0.03
& 0.62±0.03 & 0.63±0.03 & 0.62±0.03 & 0.60±0.03 & 0.61±0.03 \\

\checkmark & \checkmark
& 0.64±0.03 & 0.66±0.03 & 0.65±0.03 & 0.64±0.03 & 0.61±0.03
& 0.63±0.03 & 0.66±0.03 & 0.67±0.03 & 0.69±0.03 & 0.68±0.03 \\
\midrule

\multicolumn{12}{c}{\textbf{OGBG-molbbp} (ROC-AUC)} \\
\midrule

\xmark & \xmark
& 0.57±0.04 & 0.57±0.05 & 0.58±0.04 & 0.58±0.05 & 0.58±0.06
& 0.58±0.05 & 0.58±0.06 & 0.59±0.07 & 0.59±0.07 & 0.59±0.08 \\

\checkmark & \xmark
& 0.59±0.04 & 0.59±0.04 & 0.60±0.05 & 0.60±0.05 & 0.60±0.06
& 0.60±0.06 
& 0.57±0.10
& 0.60±0.08 & 0.60±0.08 & 0.60±0.09 \\

\xmark & \checkmark
& 0.61±0.03 & 0.61±0.03 & 0.61±0.04 & 0.61±0.05 & 0.61±0.06
& 0.61±0.07 & 0.61±0.07 & 0.61±0.08 & 0.61±0.09 & 0.61±0.09 \\

\checkmark & \checkmark
& 0.62±0.02 & 0.62±0.04 & 0.62±0.05 & 0.62±0.06 & 0.62±0.07
& 0.58±0.15
& 0.62±0.08 & 0.63±0.10 & 0.63±0.11 & 0.63±0.12 \\
\bottomrule
\end{tabular}
}
\end{table}

\begin{table}[h]
\vskip .1in
\centering
\caption{Performance comparison of DosCond across multiple datasets and percentages of graphs sampled. Performance is reported as mean ± standard deviation of accuracy (ACC) and area under the receiver operating curve (ROC-AUC), for configurations of DosCond with and without labels (lab.) and validation (val.).}
\vspace{.1in}
\resizebox{1.0\textwidth}{!}{%
\begin{tabular}{cc cccccccccc}
\toprule
\multicolumn{2}{c}{} & \multicolumn{10}{c}{\textbf{\% of graphs sampled}} \\
\cmidrule(lr){3-12}
\textbf{lab.} & \textbf{val.} 
& \textbf{1} & \textbf{2} & \textbf{3} & \textbf{4} & \textbf{5} 
& \textbf{6} & \textbf{7} & \textbf{8} & \textbf{9} & \textbf{10} \\
\midrule
\multicolumn{12}{c}{\textbf{MUTAG} (ACC)} \\
\midrule

\xmark & \xmark
& 0.671±0.003 & 0.736±0.027 & 0.731±0.008 & 0.727±0.000 & 0.709±0.008
& 0.702±0.008 & 0.696±0.021 & 0.680±0.009 & 0.716±0.006 & 0.704±0.011 \\

\checkmark & \xmark
& 0.682±0.015 & 0.730±0.020 & 0.726±0.018 & 0.719±0.016 & 0.698±0.018
& 0.691±0.021 & 0.684±0.020 & 0.679±0.015 & 0.712±0.015 & 0.705±0.020 \\

\xmark & \checkmark
& 0.656±0.035 & 0.687±0.009 & 0.687±0.024 & 0.698±0.018 & 0.667±0.022
& 0.698±0.016 & 0.700±0.020 & 0.707±0.019 & 0.704±0.017 & 0.684±0.023 \\

\checkmark & \checkmark
& 0.707±0.020 & 0.693±0.005 & 0.687±0.025 & 0.704±0.027 & 0.667±0.020
& 0.691±0.028 & 0.700±0.016 & 0.704±0.017 & 0.702±0.019 & 0.693±0.036 \\
\midrule

\multicolumn{12}{c}{\textbf{COX2} (ACC)} \\
\midrule

\xmark & \xmark
& 0.439±0.142 & 0.217±0.004 & 0.248±0.031 & 0.254±0.025 & 0.772±0.017
& 0.394±0.077 & 0.650±0.090 & 0.219±0.003 & 0.216±0.003 & 0.215±0.000 \\

\checkmark & \xmark
& 0.600±0.060 & 0.456±0.033 & 0.500±0.050 & 0.600±0.044 & 0.700±0.033
& 0.650±0.022 & 0.720±0.040 & 0.500±0.056 & 0.550±0.030 & 0.750±0.022 \\

\xmark & \checkmark
& 0.754±0.021 & 0.760±0.009 & 0.678±0.080 & 0.678±0.083 & 0.755±0.034
& 0.776±0.014 & 0.786±0.000 & 0.763±0.028 & 0.765±0.022 & 0.783±0.002 \\

\checkmark & \checkmark
& 0.744±0.051 & 0.769±0.024 & 0.783±0.004 & 0.787±0.005 & 0.780±0.006
& 0.784±0.003 & 0.780±0.016 & 0.759±0.034 & 0.784±0.005 & 0.784±0.007 \\
\midrule

\multicolumn{12}{c}{\textbf{NCI1} (ACC)} \\
\midrule
%---- (no val, no labels)
\xmark & \xmark
& 0.503±0.022 & 0.510±0.025 & 0.490±0.030
& 0.520±0.028 & 0.521±0.021
& 0.535±0.019 & 0.537±0.025 & 0.540±0.022 
& 0.512±0.040 
& 0.545±0.026 \\
%---- (no val, labels)
\checkmark & \xmark
& 0.521±0.031 & 0.530±0.028 & 0.570±0.022 & 0.512±0.039
& 0.555±0.021
& 0.560±0.033 & 0.526±0.040
& 0.575±0.020 & 0.576±0.019 & 0.580±0.022 \\

\xmark & \checkmark
& 0.540±0.020 & 0.550±0.025 & 0.555±0.019 & 0.557±0.021 
& 0.518±0.035
& 0.560±0.018 & 0.563±0.024 & 0.580±0.020 & 0.578±0.026 & 0.590±0.022 \\

\checkmark & \checkmark
& 0.562±0.012 & 0.560±0.015 
& 0.570±0.018 & 0.575±0.021 & 0.583±0.020
& 0.555±0.030
& 0.590±0.024 & 0.605±0.019 & 0.608±0.022 & 0.612±0.018 \\
\midrule

\multicolumn{12}{c}{\textbf{PROTEINS} (ACC)} \\
\midrule

\xmark & \xmark
& 0.481±0.033 & 0.670±0.009 & 0.651±0.010 & 0.663±0.004 & 0.551±0.016
& 0.634±0.003 & 0.610±0.013 & 0.645±0.006 & 0.610±0.006 & 0.628±0.009 \\

\checkmark & \xmark
& 0.620±0.010 & 0.630±0.020 & 0.645±0.025 & 0.660±0.018 & 0.583±0.012
& 0.633±0.017 & 0.650±0.020 & 0.655±0.012 & 0.650±0.023 & 0.645±0.015 \\

\xmark & \checkmark
& 0.647±0.010 & 0.653±0.009 & 0.647±0.017 & 0.668±0.003 & 0.658±0.020
& 0.641±0.002 & 0.669±0.005 & 0.647±0.016 & 0.642±0.023 & 0.663±0.007 \\

\checkmark & \checkmark
& 0.661±0.011 & 0.643±0.033 & 0.633±0.003 & 0.681±0.027 & 0.655±0.004
& 0.637±0.011 & 0.653±0.003 & 0.658±0.006 & 0.653±0.003 & 0.664±0.002 \\
\midrule

\multicolumn{12}{c}{\textbf{OGBG-MOLBACE} (ROC-AUC)} \\
\midrule

\xmark & \xmark
& 0.53±0.02 & 0.45±0.04 & 0.64±0.03 & 0.60±0.02 & 0.37±0.01
& 0.56±0.06 & 0.62±0.02 & 0.54±0.01 & 0.60±0.03 & 0.62±0.03 \\

\checkmark & \xmark
& 0.62±0.02 & 0.61±0.02 & 0.65±0.01 & 0.65±0.03 & 0.60±0.03
& 0.64±0.02 & 0.63±0.02 & 0.65±0.03 & 0.64±0.02 & 0.66±0.02 \\
\xmark & \checkmark
& 0.68±0.03 & 0.66±0.02 & 0.64±0.01 & 0.64±0.03 & 0.62±0.02
& 0.67±0.01 & 0.65±0.02 & 0.65±0.03 & 0.66±0.01 & 0.65±0.04 \\

\checkmark & \checkmark
& 0.67±0.01 & 0.67±0.03 & 0.64±0.01 & 0.67±0.01 & 0.65±0.03
& 0.68±0.02 & 0.66±0.02 & 0.66±0.01 & 0.67±0.04 & 0.66±0.02 \\
\midrule

\multicolumn{12}{c}{\textbf{OGBG-MOLBBBP} (ROC-AUC)} \\
\midrule

\xmark & \xmark
& 0.43±0.02 & 0.44±0.02 & 0.30±0.02
& 0.45±0.02 & 0.46±0.03
& 0.46±0.03 & 0.32±0.01 
& 0.47±0.03 & 0.47±0.04 & 0.48±0.03 \\

\checkmark & \xmark
& 0.45±0.03 & 0.46±0.03 & 0.46±0.03 & 0.40±0.02 
& 0.46±0.03
& 0.47±0.03 & 0.48±0.03 & 0.44±0.03 
& 0.48±0.03 & 0.49±0.03 \\

\xmark & \checkmark
& 0.48±0.02 & 0.48±0.02 & 0.49±0.02 & 0.49±0.02
& 0.50±0.03 & 0.47±0.03 
& 0.50±0.02 & 0.50±0.02 & 0.51±0.02 & 0.51±0.02 \\

\checkmark & \checkmark
& 0.51±0.01 & 0.51±0.02
& 0.52±0.02 & 0.52±0.02 
& 0.40±0.02 
& 0.54±0.02 & 0.55±0.02 & 0.56±0.02 & 0.58±0.02 & 0.62±0.02 \\
\bottomrule
\end{tabular}%
}
\end{table}

\subsection{Performance of MIRAGE}\label{sec:additional-background-experiments-mirage}

Table~\ref{tab:mirage} reports the performance of Mirage \citep{mirage} on NCI1 and OGBG-MOLBACE for certain subsample percentages, which were the only cases where we were able to execute the publicly available referenced in the original paper. Unresolved errors were encountered when running Mirage on MUTAG, PROTEINS, NCI1, DD, OGBG-MOLBBBP, and the omitted percentages in Table~\ref{tab:mirage}, due to the MP Tree search either returning an empty selection set or generating trees in a format incompatible with GNN training.

Unlike our method, MIRAGE is not label-agnostic, as it explicitly relies on a labeled validation set with an 80\%-train / 10\%-validation / 10\%-test split. Despite this supervision, MIRAGE underperforms compared to TMD and other methods, typically achieving accuracy below 0.5 for NCI1 and ROC-AUC below 0.5 for OGBG-MOLBBBP.

\begin{table}[ht]
\centering

\caption{Performance of Mirage on the NCI1 and molbbp datasets across percentages of graphs sampled, using the original implementation with labels and validation. Performance is reported as mean ± standard deviation of accuracy (ACC) and area under the receiver operating curve (ROC-AUC). Results for '--', other datasets, and configurations without labels or without validation could not be generated due to unresolved errors in execution.}
\vskip .1in
\label{tab:doscond}
\resizebox{1.0\textwidth}{!}{%
\begin{tabular}{cc cccccccccc}
\toprule
\multicolumn{1}{c}{} & \multicolumn{10}{c}{\textbf{\% of graphs sampled}} \\
\cmidrule(lr){2-11}

\textbf{Dataset} & \textbf{1} & \textbf{2} & \textbf{3} & \textbf{4} & \textbf{5} 
& \textbf{6} & \textbf{7} & \textbf{8} & \textbf{9} & \textbf{10} \\
\midrule

\makecell{\textbf{NCI1} (ACC)} & 0.509±0.021 & 0.513±0.010 & -- & 0.492±0.015 & 0.531±0.022
& 0.488±0.030 & -- & -- & 0.488±0.022 & 0.501±0.018 \\
\midrule

\makecell{\textbf{OGBG-MOLBBBP} (ROC-AUC)} & 0.51±0.0 & 0.42±0.1 & 0.49±0.0 & 0.41±0.0 & 0.42±0.1 
& 0.58±0.0 & 0.43±0.1 & 0.21±0.0 & 0.41±0.0 & 0.37±0.2 \\
\bottomrule
\end{tabular}%
}
\label{tab:mirage}
\end{table}

%\FloatBarrier

\subsection{Demonstration of generalization to alternative GNN architecture}\label{def:generalize-architecture}

To demonstrate the effect of modifying the model architecture, we also consider the effect of increasing the size of the neural network--from 128 neurons to 256 for TUDatasets (MUTAG, PROTEINS, COX2, NCI1), and from 256 to 512 for OGBG--and modify the aggregation function to global mean pool, which is another popular choice of pooling function in the applied GNN literature. 

\paragraph{Graph subsampling} Because the graph condensation methods are significantly more time-intensive and require re-running the entire distillation process for a new model architecture, we did not run KiDD and DOSCOND on this alternative architecture. 

Our results for all of the medoids-methods as well as Random are shown in Table~\ref{tab:performance_comparison_large}. We use the same experimental setup as for Table~\ref{tab:all-results-graph-subsampling1} and \ref{tab:all-results-graph-subsampling2}, except for modifying the pooling layer and number of neurons per layer, as described above. Each entry reports the mean performance (measured in test accuracy for the TUDatasets and and test AUC-ROC for the OGBG datasets) and 95\% confidence bars across 20 trials, randomized over train/test splits as well as neural network initialization. 

Combined with Table~\ref{tab:all-results-graph-subsampling1} and \ref{tab:all-results-graph-subsampling2} in the main body, our results indicate that our method can perform well on different-sized neural networks with different aggregation functions, despite not factoring this information into the choice of the graph subsamples. 

\paragraph{Node subsampling} We use the same experimental setup as for Table~\ref{tab:combined-results-nodes}, except for modifying the pooling layer and number of neurons per layer, as previously described. 

Results are shown in Table~\ref{tab:combined-results-nodes-2}. Each entry reports the mean and 95\% confidence bars across 20 trials, randomized over train/test splits as well as neural network initialization. Combined with Table~\ref{tab:combined-results-nodes} in the main body, our results indicate that our method can perform well on different-sized neural networks with different aggregation functions, despite not factoring this information into the choice of the node subsamples.

\begin{table}[ht]
\caption{Performance comparison for different methods across the percentage of subsampled graphs and various datasets on alternative GIN archiecture. Performance for MUTAG, PROTEINS, COX2, NCI1 is reported in test-accuracy. Performance for OGBG-MOLBACE and OGBG-MOLBBP is reported in test-AUC-ROC. TMD almost always performs best or second-best among the compared methods.  Dark and light green highlight best and second best performance in each row, respectively. All performances are reported as average $\pm$ 95\% confidence bars.}
\vspace{.1in}
\centering
\resizebox{1.0\textwidth}{!}{
\scriptsize
\begin{tabular}{lccccccccccc}
\toprule
Method & 1\% & 2\% & 3\% & 4\% & 5\% & 6\% & 7\% & 8\% & 9\% & 10\% \\
\midrule
\multicolumn{11}{c}{\textbf{MUTAG}} \\
\midrule
TMD & 0.58±0.07 & \cellcolor{green!80}{0.71±0.03} & \cellcolor{green!80}{0.71±0.02} & \cellcolor{green!25}{0.71±0.04} & \cellcolor{green!80}{0.76±0.04} & \cellcolor{green!80}{0.77±0.03} & \cellcolor{green!80}{0.77±0.03} & \cellcolor{green!25}{0.74±0.06} & \cellcolor{green!80}{0.75±0.04} & \cellcolor{green!80}{0.76±0.02} \\
WL & \cellcolor{green!80}{0.67±0.05} & \cellcolor{green!80}{0.71±0.04} & \cellcolor{green!25}{0.70±0.02} & \cellcolor{green!80}{0.72±0.03} & 0.72±0.03 & \cellcolor{green!25}{0.69±0.03} & 0.69±0.04 & 0.69±0.04 & \cellcolor{green!25}{0.73±0.03} & 0.72±0.04 \\
Random & \cellcolor{green!25}{0.65±0.05} & \cellcolor{green!25}{0.65±0.06} & 0.66±0.07 & 0.67±0.03 & 0.72±0.04 & \cellcolor{green!25}{0.69±0.04} & 0.72±0.04 & 0.73±0.03 & \cellcolor{green!80}{0.75±0.04} & 0.72±0.03 \\
Feature & 0.55±0.07 & 0.62±0.08 & \cellcolor{green!80}{0.71±0.03} & 0.69±0.04 & \cellcolor{green!25}{0.74±0.04} & \cellcolor{green!80}{0.77±0.03} & \cellcolor{green!25}{0.76±0.02} & \cellcolor{green!80}{0.75±0.02} & \cellcolor{green!25}{0.73±0.03} & \cellcolor{green!25}{0.74±0.03} \\
\midrule
\multicolumn{11}{c}{\textbf{PROTEINS}} \\
\midrule
TMD & \cellcolor{green!80}{0.63±0.03} & \cellcolor{green!25}{0.64±0.04} & \cellcolor{green!80}{0.69±0.02} & \cellcolor{green!80}{0.70±0.02} & \cellcolor{green!25}{0.67±0.03} & \cellcolor{green!80}{0.69±0.01} & \cellcolor{green!25}{0.69±0.02} & \cellcolor{green!80}{0.69±0.02} & \cellcolor{green!25}{0.70±0.02} & \cellcolor{green!25}{0.70±0.02} \\
WL & \cellcolor{green!25}{0.62±0.02} & \cellcolor{green!80}{0.65±0.03} & \cellcolor{green!25}{0.68±0.02} & \cellcolor{green!25}{0.68±0.02} & \cellcolor{green!80}{0.68±0.01} & \cellcolor{green!25}{0.67±0.02} & 0.68±0.02 & \cellcolor{green!80}{0.69±0.02} & 0.66±0.04 & 0.67±0.02 \\
Random & 0.61±0.02 & \cellcolor{green!25}{0.64±0.02} & 0.64±0.02 & 0.64±0.04 & \cellcolor{green!25}{0.67±0.02} & 0.65±0.02 & \cellcolor{green!25}{0.69±0.02} & \cellcolor{green!25}{0.67±0.02} & 0.68±0.02 & 0.66±0.03 \\
Feature & \cellcolor{green!25}{0.62±0.03} & 0.60±0.04 & 0.64±0.03 & 0.65±0.04 & \cellcolor{green!80}{0.68±0.03} & \cellcolor{green!25}{0.67±0.03} & \cellcolor{green!80}{0.70±0.02} & 0.66±0.04 & \cellcolor{green!80}{0.71±0.02} & \cellcolor{green!80}{0.72±0.01} \\
\midrule
\multicolumn{11}{c}{\textbf{COX2}} \\
\midrule
TMD & \cellcolor{green!80}{0.70±0.04} & \cellcolor{green!80}{0.76±0.02} & \cellcolor{green!80}{0.75±0.03} & \cellcolor{green!80}{0.78±0.02} & \cellcolor{green!80}{0.78±0.02} & \cellcolor{green!80}{0.77±0.01} & \cellcolor{green!80}{0.77±0.01} & \cellcolor{green!80}{0.78±0.02} & \cellcolor{green!25}{0.77±0.02} & \cellcolor{green!80}{0.78±0.02} \\
WL & 0.57±0.06 & 0.70±0.04 & \cellcolor{green!80}{0.75±0.02} & \cellcolor{green!25}{0.76±0.02} & \cellcolor{green!25}{0.77±0.02} & \cellcolor{green!80}{0.77±0.02} & \cellcolor{green!80}{0.77±0.02} & \cellcolor{green!25}{0.77±0.02} & \cellcolor{green!80}{0.78±0.02} & \cellcolor{green!80}{0.78±0.01} \\
Random & \cellcolor{green!25}{0.65±0.07} & 0.70±0.04 & 0.69±0.05 & 0.68±0.06 & 0.76±0.04 & 0.74±0.04 & \cellcolor{green!25}{0.74±0.03} & 0.74±0.04 & 0.75±0.06 & \cellcolor{green!25}{0.77±0.03} \\
Feature & \cellcolor{green!25}{0.65±0.06} & \cellcolor{green!25}{0.72±0.05} & \cellcolor{green!25}{0.71±0.04} & 0.72±0.02 & \cellcolor{green!80}{0.78±0.01} & \cellcolor{green!25}{0.75±0.03} & \cellcolor{green!25}{0.74±0.04} & 0.73±0.03 & \cellcolor{green!80}{0.78±0.02} & 0.76±0.02 \\
\midrule
\multicolumn{11}{c}{\textbf{NCI1}} \\
\midrule
TMD & 0.52 ± 0.01 & \cellcolor{green!25}{0.54 ± 0.02} & 0.53 ± 0.02 & \cellcolor{green!25}{0.51 ± 0.01} & \cellcolor{green!25}{0.52 ± 0.01} & 0.54 ± 0.02 & 0.55 ± 0.02 & \cellcolor{green!25}{0.52 ± 0.01} & 0.54 ± 0.02 & 0.53 ± 0.02 \\
WL & 0.50 ± 0.01 & 0.51 ± 0.00 & 0.52 ± 0.01 & \cellcolor{green!25}{0.51 ± 0.01} & 0.51 ± 0.01 & 0.50 ± 0.00 & 0.50 ± 0.01 & 0.50 ± 0.01 & 0.52 ± 0.01 & 0.51 ± 0.01 \\
Random & \cellcolor{green!80}{0.56 ± 0.02} & \cellcolor{green!80}{0.60 ± 0.02} & \cellcolor{green!25}{0.60 ± 0.02} & \cellcolor{green!80}{0.61 ± 0.00} & \cellcolor{green!80}{0.63 ± 0.01} & \cellcolor{green!80}{0.62 ± 0.01} & \cellcolor{green!80}{0.63 ± 0.01} & \cellcolor{green!80}{0.62 ± 0.01} & \cellcolor{green!25}{0.62 ± 0.02} & \cellcolor{green!80}{0.64 ± 0.01} \\
Feature & \cellcolor{green!25}{0.53 ± 0.01} & \cellcolor{green!80}{0.60 ± 0.02} & \cellcolor{green!80}{0.61 ± 0.01} & \cellcolor{green!80}{0.61 ± 0.02} & \cellcolor{green!80}{0.63 ± 0.01} & \cellcolor{green!25}{0.60 ± 0.02} & \cellcolor{green!25}{0.62 ± 0.02} & \cellcolor{green!80}{0.62 ± 0.01} & \cellcolor{green!80}{0.63 ± 0.00} & \cellcolor{green!25}{0.62 ± 0.01} \\
\midrule
\multicolumn{11}{c}{\textbf{OGBG-MOLBACE}} \\
\midrule
TMD & 0.45±0.01 & \cellcolor{green!80}{0.52±0.03} & \cellcolor{green!80}{0.55±0.04} & 0.49±0.02 & \cellcolor{green!25}{0.54±0.03} & \cellcolor{green!80}{0.54±0.03} & \cellcolor{green!80}{0.58±0.03} & \cellcolor{green!80}{0.60±0.03} & \cellcolor{green!80}{0.57±0.02} & \cellcolor{green!80}{0.58±0.02} \\
WL & 0.48±0.01 & \cellcolor{green!25}{0.49±0.02} & 0.48±0.02 & \cellcolor{green!25}{0.51±0.03} & 0.48±0.02 & 0.50±0.02 & 0.53±0.03 & 0.50±0.03 & 0.55±0.03 & 0.53±0.03 \\
Random & \cellcolor{green!25}{0.49±0.02} & \cellcolor{green!25}{0.49±0.03} & 0.51±0.03 & 0.49±0.02 & 0.48±0.03 & 0.50±0.04 & 0.48±0.02 & 0.48±0.02 & 0.49±0.03 & 0.51±0.03 \\
Feature & \cellcolor{green!80}{0.50±0.03} & \cellcolor{green!80}{0.52±0.03} & \cellcolor{green!25}{0.53±0.02} & \cellcolor{green!80}{0.55±0.02} & \cellcolor{green!80}{0.55±0.02} & \cellcolor{green!25}{0.51±0.02} & \cellcolor{green!25}{0.55±0.02} & \cellcolor{green!25}{0.56±0.03} & \cellcolor{green!25}{0.56±0.02} & \cellcolor{green!25}{0.56±0.02} \\
\midrule
\multicolumn{11}{c}{\textbf{OGBG-MOLBBBP}} \\
\midrule
TMD & 0.69 ± 0.07 & 0.67 ± 0.07 & \cellcolor{green!25}{0.75 ± 0.02} & \cellcolor{green!80}{0.76 ± 0.00} & \cellcolor{green!80}{0.76 ± 0.01} & \cellcolor{green!80}{0.77 ± 0.02} & \cellcolor{green!25}{0.77 ± 0.01} & \cellcolor{green!25}{0.76 ± 0.01} & \cellcolor{green!25}{0.76 ± 0.01} & 0.76 ± 0.01 \\
WL & 0.44 ± 0.10 & \cellcolor{green!25}{0.74 ± 0.04} & 0.73 ± 0.05 & \cellcolor{green!25}{0.75 ± 0.03} & \cellcolor{green!80}{0.76 ± 0.01} & \cellcolor{green!25}{0.76 ± 0.01} & \cellcolor{green!80}{0.78 ± 0.01} & \cellcolor{green!25}{0.76 ± 0.01} & \cellcolor{green!80}{0.78 ± 0.01} & \cellcolor{green!80}{0.78 ± 0.00} \\
Random & \cellcolor{green!25}{0.74 ± 0.02} & 0.65 ± 0.08 & 0.68 ± 0.07 & \cellcolor{green!80}{0.76 ± 0.01} & \cellcolor{green!80}{0.76 ± 0.01} & 0.72 ± 0.07 & 0.67 ± 0.09 & 0.73 ± 0.04 & \cellcolor{green!25}{0.76 ± 0.00} & \cellcolor{green!25}{0.77 ± 0.01} \\
Feature & \cellcolor{green!80}{0.78 ± 0.01} & \cellcolor{green!80}{0.77 ± 0.01} & \cellcolor{green!80}{0.77 ± 0.01} & 0.71 ± 0.07 & \cellcolor{green!80}{0.76 ± 0.02} & 0.72 ± 0.07 & 0.76 ± 0.01 & \cellcolor{green!80}{0.77 ± 0.01} & 0.72 ± 0.07 & \cellcolor{green!80}{0.78 ± 0.00} \\
\bottomrule
\end{tabular}
\label{tab:performance_comparison_large}
}
\end{table}

\begin{table*}[t]
\scriptsize
\caption{ Performance comparison for different methods across the percentage of subsampled graphs and various datasets on alternative GIN archiecture. All performances are reported in test accuracy. TMD tends to perform better than other methods, but is comparable with RW on COX-2 and PROTEINS on this alternative architecture.  Dark and light green highlight best and second best performance in each row, respectively. All results display average $\pm$ 95\% confidence bars. }
\vspace{0.1in}
\centering

\begin{subtable}{\textwidth}
\centering
\resizebox{1.0\textwidth}{!}{%
\begin{tabular}{l|ccccccccc|c}
\hline
\textbf{Method} & \textbf{1\%} & \textbf{2\%} & \textbf{3\%} & \textbf{4\%} & \textbf{5\%} & \textbf{6\%} & \textbf{7\%} & \textbf{8\%} & \textbf{9\%} & \textbf{Wins} \\
\hline
\textbf{RW}
  & \cellcolor{green!25}{0.66±0.04} 
  & \cellcolor{green!80}{0.67±0.03} 
  & 0.66±0.03 
  & 0.66±0.04
  & 0.61±0.06
  & 0.64±0.07
  & \cellcolor{green!25}{0.72±0.03}
  & \cellcolor{green!25}{0.75±0.04}
  & 0.79±0.04
  & 1 \\
\textbf{k-cores}
  & 0.59±0.07 
  & \cellcolor{green!25}{0.63±0.04} 
  & 0.45±0.07 
  & \cellcolor{green!25}{0.69±0.03}
  & \cellcolor{green!25}{0.63±0.06}
  & 0.62±0.08
  & \cellcolor{green!25}{0.72±0.03}
  & 0.74±0.04
  & 0.74±0.05
  & 0 \\
\textbf{Random}
  & \cellcolor{green!80}{0.70±0.03}
  & \cellcolor{green!80}{0.67±0.03} 
  & \cellcolor{green!80}{0.70±0.03} 
  & \cellcolor{green!80}{0.70±0.03}
  & \cellcolor{green!80}{0.68±0.04}
  & \cellcolor{green!25}{0.67±0.04}
  & 0.71±0.03
  & \cellcolor{green!25}{0.75±0.03}
  & \cellcolor{green!25}{0.81±0.03}
  & 5 \\
\textbf{TMD}
  & 0.42±0.07 
  & \cellcolor{green!80}{0.67±0.03} 
  & \cellcolor{green!25}{0.68±0.04} 
  & 0.68±0.02
  & \cellcolor{green!80}{0.68±0.04}
  & \cellcolor{green!80}{0.74±0.03}
  & \cellcolor{green!80}{0.75±0.04}
  & \cellcolor{green!80}{0.80±0.02}
  & \cellcolor{green!80}{0.83±0.03}
  & 6 \\
\hline
\end{tabular}
}
\caption{MUTAG}
\label{tab:mutag-node-2-pivot}
\end{subtable}

\vspace{1em}

\begin{subtable}{\textwidth}
\centering
\resizebox{1.0\textwidth}{!}{%
\begin{tabular}{l|ccccccccc|c}
\hline
\textbf{Method} & \textbf{1\%} & \textbf{2\%} & \textbf{3\%} & \textbf{4\%} & \textbf{5\%} & \textbf{6\%} & \textbf{7\%} & \textbf{8\%} & \textbf{9\%} & \textbf{Wins} \\
\hline
\textbf{RW}
  & \cellcolor{green!80}{0.61±0.01}
  & \cellcolor{green!80}{0.60±0.01}
  & \cellcolor{green!80}{0.63±0.01}
  & 0.63±0.01
  & \cellcolor{green!25}{0.67±0.02}
  & \cellcolor{green!25}{0.69±0.01}
  & \cellcolor{green!25}{0.68±0.02}
  & \cellcolor{green!80}{0.72±0.01}
  & \cellcolor{green!80}{0.70±0.02}
  & 5 \\
\textbf{k-cores}
  & 0.59±0.01
  & \cellcolor{green!80}{0.60±0.01}
  & 0.60±0.01
  & \cellcolor{green!25}{0.64±0.02}
  & 0.65±0.01
  & 0.68±0.02
  & 0.68±0.02
  & \cellcolor{green!25}{0.69±0.02}
  & \cellcolor{green!25}{0.69±0.02}
  & 1 \\
\textbf{Random}
  & \cellcolor{green!25}{0.60±0.01}
  & \cellcolor{green!80}{0.60±0.01}
  & \cellcolor{green!25}{0.62±0.02}
  & \cellcolor{green!80}{0.65±0.02}
  & \cellcolor{green!25}{0.67±0.02}
  & \cellcolor{green!25}{0.69±0.02}
  & \cellcolor{green!80}{0.69±0.02}
  & 0.68±0.02
  & \cellcolor{green!80}{0.70±0.02}
  & 4 \\
\textbf{TMD}
  & \cellcolor{green!25}{0.60±0.01}
  & \cellcolor{green!80}{0.60±0.01}
  & 0.61±0.01
  & \cellcolor{green!25}{0.64±0.01}
  & \cellcolor{green!80}{0.67±0.01}
  & \cellcolor{green!80}{0.70±0.02}
  & \cellcolor{green!80}{0.70±0.02}
  & \cellcolor{green!25}{0.69±0.02}
  & \cellcolor{green!80}{0.70±0.03}
  & 5 \\
\hline
\end{tabular}
}
\caption{PROTEINS}
\label{tab:proteins-node-2-pivot}
\end{subtable}

\vspace{1em}

\begin{subtable}{\textwidth}
\centering
\resizebox{1.0\textwidth}{!}{%
\begin{tabular}{l|ccccccccc|c}
\hline
\textbf{Method} & \textbf{1\%} & \textbf{2\%} & \textbf{3\%} & \textbf{4\%} & \textbf{5\%} & \textbf{6\%} & \textbf{7\%} & \textbf{8\%} & \textbf{9\%} & \textbf{Wins} \\
\hline
\textbf{RW}
  & \cellcolor{green!25}{0.58±0.01}
  & \cellcolor{green!80}{0.59±0.01}
  & \cellcolor{green!80}{0.61±0.01}
  & \cellcolor{green!25}{0.64±0.02}
  & \cellcolor{green!80}{0.67±0.02}
  & 0.65±0.02
  & \cellcolor{green!25}{0.72±0.02}
  & \cellcolor{green!25}{0.73±0.02}
  & 0.73±0.03
  & 3 \\
\textbf{k-cores}
  & \cellcolor{green!80}{0.59±0.01}
  & \cellcolor{green!80}{0.59±0.01}
  & 0.59±0.02
  & 0.59±0.01
  & 0.60±0.01
  & 0.59±0.01
  & 0.59±0.01
  & 0.60±0.01
  & 0.59±0.01
  & 2 \\
\textbf{Random}
  & \cellcolor{green!80}{0.59±0.01}
  & \cellcolor{green!25}{0.58±0.01}
  & \cellcolor{green!25}{0.60±0.01}
  & \cellcolor{green!25}{0.64±0.02}
  & \cellcolor{green!25}{0.66±0.02}
  & \cellcolor{green!80}{0.69±0.02}
  & 0.71±0.02
  & \cellcolor{green!80}{0.74±0.01}
  & \cellcolor{green!25}{0.74±0.01}
  & 3 \\
\textbf{TMD}
  & \cellcolor{green!25}{0.58±0.01}
  & \cellcolor{green!80}{0.59±0.01}
  & \cellcolor{green!25}{0.60±0.01}
  & \cellcolor{green!80}{0.65±0.01}
  & 0.65±0.01
  & \cellcolor{green!25}{0.67±0.03}
  & \cellcolor{green!80}{0.73±0.02}
  & 0.71±0.03
  & \cellcolor{green!80}{0.75±0.02}
  & 4 \\
\hline
\end{tabular}
}
\caption{DD}
\label{tab:dd-node-2-pivot}
\end{subtable}

\vspace{1em}

\begin{subtable}{\textwidth}
\centering
\resizebox{1.0\textwidth}{!}{%
\begin{tabular}{l|ccccccccc|c}
\hline
\textbf{Method} & \textbf{1\%} & \textbf{2\%} & \textbf{3\%} & \textbf{4\%} & \textbf{5\%} & \textbf{6\%} & \textbf{7\%} & \textbf{8\%} & \textbf{9\%} & \textbf{Wins} \\
\hline
\textbf{RW}
  & \cellcolor{green!25}{0.78±0.01}
  & 0.77±0.01
  & \cellcolor{green!80}{0.79±0.01}
  & \cellcolor{green!80}{0.80±0.02}
  & \cellcolor{green!80}{0.79±0.02}
  & \cellcolor{green!80}{0.78±0.02}
  & \cellcolor{green!25}{0.79±0.02}
  & \cellcolor{green!80}{0.79±0.02}
  & 0.77±0.02
  & 5 \\
\textbf{k-cores}
  & \cellcolor{green!25}{0.78±0.01}
  & \cellcolor{green!25}{0.78±0.02}
  & 0.77±0.01
  & \cellcolor{green!25}{0.79±0.02}
  & 0.74±0.05
  & \cellcolor{green!80}{0.78±0.01}
  & 0.77±0.02
  & \cellcolor{green!25}{0.78±0.02}
  & \cellcolor{green!80}{0.79±0.02}
  & 2 \\
\textbf{Random}
  & \cellcolor{green!80}{0.79±0.02}
  & \cellcolor{green!25}{0.78±0.02}
  & \cellcolor{green!25}{0.78±0.01}
  & \cellcolor{green!25}{0.79±0.02}
  & \cellcolor{green!80}{0.79±0.01}
  & \cellcolor{green!80}{0.78±0.02}
  & 0.77±0.02
  & \cellcolor{green!25}{0.78±0.01}
  & \cellcolor{green!25}{0.78±0.01}
  & 3 \\
\textbf{TMD}
  & \cellcolor{green!80}{0.79±0.02}
  & \cellcolor{green!80}{0.79±0.02}
  & \cellcolor{green!25}{0.78±0.01}
  & 0.77±0.02
  & \cellcolor{green!25}{0.78±0.01}
  & \cellcolor{green!80}{0.78±0.02}
  & \cellcolor{green!80}{0.80±0.02}
  & \cellcolor{green!25}{0.78±0.03}
  & \cellcolor{green!80}{0.79±0.02}
  & 5 \\
\hline
\end{tabular}
}
\caption{COX2}
\label{tab:cox2-node-2-pivot}
\end{subtable}
\label{tab:combined-results-nodes-2}
\end{table*}

\FloatBarrier
\newpage
\section{Omitted Proofs}\label{sec:apx}

In this Appendix, we include full proofs of the theoretical results, which formally expand the proof sketches in the main body. Throughout this section, we use $\text{deg}_G(v)$ to denote the degree of $v \in V$. 

\subsection{Omitted proofs from Section~\ref{sec:graph_drop}}\label{sec:apx_graph_drop}

\tmdgen*
\begin{proof} Let $\kappa: [n] \to \cI$ map $i \in [n]$ to the index $j \in \cI$ such that $G_j$ is the closest graph in $\{G_i\}_{i \in \cI}$ to $G_i.$ That is, 
\begin{align*}
    \kappa(i) = \argmin_{j \in \cI} \cTMD_w^L(G_i, G_j), 
\end{align*}
with ties broken arbitrarily (but consistently with the mapping $\tau$.) First, we can rearrange the sums as follows
\begin{align*}
    \abs{\frac{1}{n} \sum_{i \in \cI} \tau_i \cL(h(G_i); y_i) - \frac{1}{n}\sum_{i \in [n]}\cL(h(G_i); y_i)} &= \abs{ \frac{1}{n}\sum_{i \in [n]} \cL(h(G_{\kappa(i)}); y_{\kappa(i)})- \cL(h(G_i); y_i)} \\
    &\leq  \frac{1}{n}\sum_{i \in [n]} \abs{\cL(h(G_{\kappa(i)}; y_{\kappa(i)}))- \cL(h(G_i); y_i)} \\
    &\leq  \frac{M}{n}\sum_{i \in [n]} \norm{\cL(h(G_{\kappa(i)}); y_{\kappa(i)})- \cL(h(G_i); y_i)},
\end{align*}
where in the last inequality, we used that $\cL$ is $M$-lipschitz. Now, by Theorem~\ref{thm:stable}, we have 
\begin{align*}
    \frac{M}{n}\sum_{i \in [n]} \norm{\cL(h(G_{\kappa(i)}); y_{\kappa(i)})- h(G_i; y_i)} &\leq \frac{M}{n} \paren{\prod_{\ell \in [L-1]} \Phi_\ell} \sum_{i \in [n]} \cTMD_w^L(G_{\kappa(i)}, G_i) \\
    &= M \paren{\prod_{\ell \in [L-1]} \Phi_\ell} f_{\cTMD_w^L}(S; X).
\end{align*}
\end{proof}

\erm*
\begin{proof}
From Lemma~\ref{lemma:tmd-gen}, we know that \[\frac{1}{n}\sum_{i \in [n]}\cL\paren{\hat{h}(G_i); y_i} \leq \frac{1}{n} \sum_{i \in S} \tau_{i} \cL(\hat{h}(G_i); y_i) + c\epsilon.\]
    Let $h^*$ be the hypothesis that minimizes average loss over the original dataset $X$: \[h^* = \argmin_{h \in \cH}\frac{1}{n}\sum_{i \in [n]}\cL(h(G_i); y_i).\] By definition of $\hat{h}$, this means that \[\frac{1}{n}\sum_{i \in [n]}\cL\paren{\hat{h}(G_i); y_i} \leq \frac{1}{n} \sum_{i \in S} \tau_{i} \cL(h^*(G_i); y_i) + c\epsilon.\] Applying Lemma~\ref{lemma:tmd-gen} again, we have that  \[\frac{1}{n}\sum_{i \in [n]}\cL\paren{\hat{h}(G_i); y_i} \leq \frac{1}{n}\sum_{i \in [n]}\cL\paren{h^*(G_i); y_i} + 2c\epsilon.\]
\end{proof}

\subsubsection{Proof of Theorem~\ref{thm:TMD-is-needed}}\label{sec:thm-tmd-is-needed}

In this section, we prove Theorem~\ref{thm:TMD-is-needed}. The key insight driving our analysis is that in order to disprove Conjecture~\ref{conj:TMD-not-needed} for a specific graph pseudo-metric $D$, it suffices to produce a pair of graphs $G, G'$ and an $L$-layer GNN $h$ such that $D(G, G') = 0$ while $h(G) \neq h(G')$. In words, this means that if we have a GNN that outputs different values on two graphs $G, G'$ (i..e, the GNN can \emph{distinguish} them) but $D(G, G') = 0$, then we \emph{cannot hope} to show that the difference in GNN outputs is bounded proportional to $D(G, G')$.
The following Lemma formalizes this intuition.

\begin{lemma}\label{lemma:sufficient-condition} Let $D$ denote a graph pseudo-metric. Suppose there exists a pair of graphs $G, G'$ and a GNN  $h: \cG \to \R^d$ with $\ell$-th layer Lipschitz constant $\Phi_\ell$ such that $D(G, G') = 0$ and $h(G) \neq h(G')$. Then, Conjecture~\ref{conj:TMD-not-needed} is not true for the kernel $D$. 
\end{lemma}
\begin{proof} Suppose for the sake of contradiction that Conjecture~\ref{conj:TMD-not-needed} is true for the kernel $D$. Then, there is a constant $C(\{\phi_\ell\}_{\ell=1}^{L}) > 0$ (depending on the Lipschitz-constants $\{\phi_\ell\}_{\ell=1}^{L}$) such that $\norm{h(G) - h(G')} \leq C(\{\phi_\ell\}_{\ell=1}^{L}) \cdot D(G, G') = 0$. Since norms are positive definite, this contradicts that $h(G) \neq h(G')$. 
\end{proof}

In the following lemmas, we show that the assumption of Lemma~\ref{lemma:sufficient-condition} holds for four of the most commonly used graph pseudo-metrics in the graph learning literature: WL (Weisfeller-Lehman), WL-OA (Weisfeller-Lehman Optimal Transport), SP (Shortest Paths), and GS (Graphlet Sampling.) Indeed, the fact that the assumption of Lemma~\ref{lemma:sufficient-condition} holds for Shortest Paths and Graphlet Sampling was previously shown by \citet{kriege2020survey} (Figure 7 of \citet{kriege2020survey}), in which the authors demonstrated examples of graphs where the Shortest Paths and Graphlet Sampling pseudo-metrics would equal 0 \emph{even} when a GNN can distinguish between them. 

\begin{lemma}[Result of \citet{kriege2020survey}]\label{lemma:partial} The assumption of Lemma~\ref{lemma:sufficient-condition} holds for $D = D_{GS}$ and $D= D_{SP}$. 
\end{lemma}

However, such a result (to our knowledge) was not previously known for the original WL (Weisfeller-Lehman) pseudo-metric or its variant WL-OA (Weisfeller-Lehman Optimal Transport) 
 We refer to these metrics as WL-based metrics. To prove that the assumption of Lemma~\ref{lemma:sufficient-condition} holds for these kernels, we first give an overview of how the WL-based metrics are constructed. First, we define WL iterations as follows \citep{zhang2017weisfeiler, shervashidze2011weisfeiler}. These form the basis of the definition of the various WL-based pseudo-metrics. 

\begin{definition}[WL Iterations \citep{zhang2017weisfeiler}]\label{def:WL-iterations}
  Let \(G = (V, E)\) be a labeled graph with initial labels \(\ell_0(v)\) for each \(v \in V\).
  Let \(\Gamma(v)\) denote the set of neighbors of \(v\).
  For a fixed number of iterations \(L \ge 0\), the \emph{Weisfeiler--Lehman (WL) labeling} proceeds as follows:

  \begin{itemize}
    \item For each iteration \(h = 1, 2, \dots, L\):
    \begin{enumerate}
      \item \textbf{Multiset labeling:} 
        \[
          m_h(v) \;=\; \{\ell_{h-1}(u) : u \in \Gamma(v)\}.
        \]
      \item \textbf{Label compression:} 
        \[
          \ell_h^G(v) \;=\; \mathrm{Hash}\!\bigl(\ell_{h-1}(v),\, m_h(v)\bigr),
        \]
        where \(\mathrm{Hash}\) is an injective function (often a hash or string-encoding).
      \item \textbf{Feature extraction:} For each iteration \(h\),
        define a feature map \(\phi^{(h)}(G)\in\mathbb{R}^{|\mathcal{L}_h|}\), 
        counting the occurrences of each \emph{compressed} label in \(G\) at iteration \(h\).
        Here, \(\mathcal{L}_h\) is the set of all possible labels at iteration \(h\).
    \end{enumerate}
  \end{itemize}

We call \(\{\ell_h^G\}_{h=0}^L\) the WL-refined labels of \(G\), and
  \(\{\phi^{(h)}(G)\}_{h=0}^L\) the corresponding feature vectors (for each iteration $\ell \in [L])$.
\end{definition}

Equipped with this definition, we now define the WL kernel and pseudo-metric. 

\begin{definition}[WL \citep{shervashidze2011weisfeiler}]\label{def:WL-kernel}
  Given two graphs \(G\) and \(G'\) and an integer $L > 0$, let \(\{\phi^{(h)}(G)\}_{h=0}^L\) and \(\{\phi^{(h)}(G')\}_{h=0}^L\) be their WL feature vectors from Definition~\ref{def:WL-iterations}.
  The \emph{Weisfeiler--Lehman (WL) kernel} is defined by
  \[
    k_{\mathrm{WL}}(G, G') 
    \;=\; 
    \sum_{h=0}^L
      \bigl\langle \phi^{(h)}(G),\, \phi^{(h)}(G') \bigr\rangle,
  \]
  where \(\langle \cdot,\cdot\rangle\) is the standard dot product in \(\mathbb{R}^{|\mathcal{L}_h|}\). From the WL kernel \(k^L_{\mathrm{WL}}\), one can define a \emph{pseudo-metric} 
  \(D_{\mathrm{WL}}(G, G')\) by
  \[
    D^L_{\mathrm{WL}}(G, G')
    \;=\;
    \sqrt{
      k_{\mathrm{WL}}(G, G)
      \;+\;
      k_{\mathrm{WL}}(G', G')
      \;-\;
      2\,k_{\mathrm{WL}}(G, G')
    }.
  \]
\end{definition}

The WL-OA kernel is defined similarly; however, we perform an optimal assignment between the WL- labels. 

\begin{definition}[WL-OA Kernel]\label{def:WL-OA-kernel}
  After performing \(L\) WL iterations on a graph \(G\), each node \(v \in V(G)\) 
  has a final label (or embedding) \(\ell_L(v)\).  
  Let \(\kappa\bigl(\ell_L(v), \ell_L(w)\bigr) = \begin{cases}
      1, & \ell_L(v) = \ell_L(w) \\
      0, & \text{otherwise}.
  \end{cases}\) An \emph{optimal assignment} \(\alpha : V(G) \to V(G')\) between two graphs \(G\) 
  and \(G'\) maximizes the total node-level similarity:
  \[
    \max_{\alpha : V(G)\!\to\!V(G')}
    \sum_{v \in V(G)}
      \kappa\!\bigl(\ell_L(v),\, \ell_L(\alpha(v))\bigr).
  \] The WL-OA kernel is then defined as follows:
  \[
    k_{\mathrm{WL\text{-}OA}}(G, G')
    \;=\;
    \sum_{h=0}^L\;
      \max_{\alpha_h : V(G)\!\to\!V(G')} 
      \sum_{v \in V(G)} 
        \kappa\bigl(\ell_h(v), \ell_h(\alpha_h(v))\bigr).
  \]
  From the WL-OA kernel \(k^L_{\mathrm{WL-OA}}\), one can define a \emph{pseudo-metric} 
  \(D_{\mathrm{WL-OA}}(G, G')\) by
  \[
    D^L_{\mathrm{WL-OA}}(G, G')
    \;=\;
    \sqrt{
      k_{\mathrm{WL-OA}}(G, G)
      \;+\;
      k_{\mathrm{WL-OA}}(G', G')
      \;-\;
      2\,k_{\mathrm{WL-OA}}(G, G')
    }.
  \]
\end{definition}

Next, we construct a pair of graphs which cannot be distinguished by the WL or WL-OA kernels. 
\begin{lemma}\label{lemma:prev} Let $G = (V, E, f)$ and $G' = (V, E, f')$ where $V = [n]$, $f_i = i$ for all $i \in [n]$, and $f'_i = 10\cdot i$ for all $i \in [n]$. Then, $D^L_{\mathrm{WL-OA}}(G, G') = D^L_{\mathrm{WL}}(G, G') = 0$ for every integer $L > 0$. 
\end{lemma}
\begin{proof} Note that $G, G'$ have the exact same graph structure (i.e., edges) and differ only in the feature attributes $f, f'$ respectively. Consequently, by Definition~\ref{def:WL-iterations}, we see that $\{\ell_h^G(i)\}_{h=0}^L = \{\ell_h^{G'(i)}\}_{h=0}^L$ for all $i \in [n]$. Since the distance pseudo-metrics $D^L_{\mathrm{WL}}$ and $D^L_{\mathrm{WL-OA}}$ depend on $G, G'$ only through these label functions, the lemma follows.
\end{proof}

Since a GNN can trivially distinguish between these two graphs proposed in Lemma~\ref{lemma:prev}, we can prove that the assumption of Lemma~\ref{lemma:sufficient-condition} holds for the WL and WL-OA pseudo-metrics. 

\begin{corollary}\label{corr:WL}The assumption of Lemma~\ref{lemma:sufficient-condition} holds for $D = D^L_{WL}$ and $D = D^L_{WL-OA}$. 
\end{corollary}
\begin{proof} A single-layer graph neural network can distinguish between the candidates $G, G'$ guaranteed by Lemma~\ref{lemma:prev}, by simply setting all aggregation functions and nonlinearities to identity mappings. Thus, the assumption of Lemma~\ref{lemma:sufficient-condition} holds for $D = D^L_{WL}$ and $D = D^L_{WL-OA}$. 
\end{proof}

\tmdneeded*
\begin{proof} Combining Lemma~\ref{lemma:sufficient-condition} with Lemma~\ref{lemma:partial} proves the result for $D \in \{D_{SP}, D_{GS}\}$.  Combining Lemma~\ref{lemma:sufficient-condition} with Corollary~\ref{corr:WL} proves the result for $D \in \{D_{WL}, D_{WL-OA}\}$. 
\end{proof}

\subsection{Omitted proofs from Section~\ref{sec:node_drop}}\label{sec:apx-node}

\corrermfirst*
\begin{proof} Consider any $h \in \cH$. By Theorem~\ref{thm:stable}, we have that for each $i \in [n]$, 
\begin{align}\label{eq:stability-eq}
    \norm{{h}(G_i; y_i) - {h}(G_i[S_i]; y_i)} \leq \paren{\prod_{\ell \in [L-1]} \Phi_\ell } \cTMD_w^L(G_i, G_i[S_i]). 
\end{align}
Consequently, using the fact that $\cL$ is $M$-Lipschitz and \eqref{eq:stability-eq}, 
\begin{align*}
    \abs{\cL\paren{{h}(G_i); y_i} - \cL\paren{{h}(G_i[S_i])}} &\leq M \norm{{h}(G_i; y_i) - {h}(G_i[S_i]; y_i)} \\
    &\leq M \paren{\prod_{\ell \in [L-1]} \Phi_\ell } \cTMD_w^L(G_i, G_i[S_i]) \\
    &= c \cTMD_w^L(G_i, G_i[S_i]) \leq c\epsilon_i. 
\end{align*}
Consequently, by the triangle inequality, 
\begin{align*}
    \abs{ \frac{1}{n}\sum_{i \in [n]} \cL\paren{{h}(G_i); y_i} - \frac{1}{n}\sum_{i \in [n]} \cL\paren{{h}(G_i[S_i])}} &\leq \frac{1}{n}\sum_{i \in [n]} \abs{\cL\paren{{h}(G_i); y_i} - \cL\paren{{h}(G_i[S_i]; y_i)}} \\
    &\leq \frac{c}{n}\sum_{i \in [n]} \epsilon_i. 
\end{align*}
So, we know that for any $h \in \cH$
\begin{align}\label{eq:gen}
    \frac{1}{n} \sum_{i \in [n]} \cL({h}(G_i; y_i) \leq \frac{1}{n} \sum_{i \in [n]} \cL(h({G_i[S_i]}; y_i) + \frac{c}{n} \sum_{i\in[n]} \epsilon_i.
\end{align}
Let $h^\star$ be the hypothesis that minimizes average loss across the original dataset $X$: 
\begin{align*}
    h^\star = \argmin_{h \in \cH} \frac{1}{n} \sum_{i \in [n]} \cL(h(G_i); y_i). 
\end{align*}
By definition of $\hat{h}$, this means that
\begin{align*}
    \frac{1}{n} \sum_{i \in [n]} \cL(\hat{h}(G_i[S_i]; y_i) \leq \frac{1}{n} \sum_{i \in [n]} \cL(h^\star(G_i[S_i]); y_i) + \frac{c}{n} \sum_{i \in [n]} \epsilon_i. 
\end{align*}
Applying \eqref{eq:gen} again to $h^\star$, we have that 
\begin{align*}
    \frac{1}{n} \sum_{i \in [n]} \cL(\hat{h}(G_i[S_i]; y_i) \leq  \frac{1}{n} \sum_{i \in [n]} \cL(h^\star(G_i); y_i) + \frac{c}{n} \sum_{i \in [n]} \epsilon_i. 
\end{align*}
\end{proof}

\generaldecomposition*
\begin{proof} Let $\bar{G}[S]$ be the graph obtained by augmenting $G[S]$ with node features $\bm{0}$ for each $v \in V \setminus S.$ That is, $\bar{G}[S] = (V, E[S], {f'})$ where $f'^v = f^v$ if $v \in S$ and $\bm{0}$ otherwise By the definition of the augmentation function $\rho$ and $\cOTbar$ (recall \eqref{eq:rho}, \eqref{eq:OT-bar}), we can instead consider the following OT problem, which is equivalent to \eqref{eq:ot-problem}:
\begin{align*}
    \cOT_{\cTD_w}\paren{\cT_G^L, \cT_{\bar{G}[S]}^L}. 
\end{align*}

In the base case where $L=1$, we see that \emph{any} transportation plan between the nodes of $\bar{G}[S]$ and $G$ must transport the excess node feature mass $\sum_{v \notin S} \norm{f^v}$ in $G$ to some nodes in $G[S].$ The transportation plan $\bm{I}$ has cost exactly equal to $\sum_{v \in V} \norm{f^v - f'^v} = \sum_{v \notin S} \norm{f^v}$, so $\bm{I}$ must be an optimal transportation plan. 

Now, if $L > 1$, consider the OT problem~\ref{eq:ot-problem}, and let $C$ and $\Gamma$ be the distance matrix and feasible transportation plans for this OT problem (recall Definition~\ref{def:OT}). Let $\gamma \in \Gamma$ be any feasible transportation plan.

By the definition of $\cTD$, for any $i, j \in V$, transporting $\gamma_{i,j}$ mass from $i$ to $j$ incurs two costs:
\begin{enumerate}[label={(\arabic*)}]
    \item $\gamma_{i,j}$ incurs the cost of transporting $f^i$ to $f'^j$.
    \item $\gamma_{i,j}$ \emph{also} incurs the cost of taking all of the depth $(L-1)$ computation trees of all neighbors of $i$ in $T_i^L(G)$, and transporting them to the depth $(L-1)$ computation trees of all neighbors of $j$ in $T_j^L(G[S])$ (after appropriately augmenting with isolated nodes of feature $\bm{0}$ for each $v \notin S$).
\end{enumerate}
To quantify these two costs, let us define $\bar{\cT}_j(T_j^L(G[S]))$ to be the multiset of computation trees obtained by augmenting $\cT_j(T_j^L(G[S]))$ with isolated nodes of feature vectors $\bm{0}$ for each $j \notin S$. That is, let $\bar{\cT}_j(T_j^L(G[S]))$ be defined as the second output of $\rho$ (recall \eqref{eq:rho}) when applied to $\paren{\cT_j(T_j^L(G)), \cT_j(T_j^L(G[S]))}:$
\begin{align*}
\paren{\cT_j(T_j^L(G)), \bar{\cT}_j(T_j^L(G[S]))} \defeq \rho\paren{\cT_j(T_j^L(G)), \cT_j(T_j^L(G[S]))}.
\end{align*}
We can now quantify the cost of transportation plan $\gamma$ in \eqref{eq:ot-problem} as follows. In the following equation, the first term corresponds to item (1) above and the second term corresponds to item (2) discussed above. 
\begin{align*}
    \sum_{i,j \in V} \gamma_{i,j} C_{i,j} &= \sum_{i,j \in V} \gamma_{i,j} \norm{f^i - f'^j} + w(L-1) \sum_{i,j \in V} \gamma_{i,j} \cOT_{\cTD_{w}}{(\cT_i(T_i^L(G)), \bar{\cT}_j(T_j^L({G}[S])))}. 
\end{align*}
Now, we can immediately lower bound the cost $\sum_{i,j} \gamma_{i,j} C_{i,j}$ of $\gamma$ as follows. Using the fact that the minimum of the sum of two functions is at least as large as the sum of their minimums, we have
\begin{align*}
    \sum_{i,j \in V} \gamma_{i,j} C_{i,j} &\geq \min_{\nu \in \Gamma}\sum_{i,j\in V} \nu_{i,j} \norm{f^i - f'^j} \\
    &+ w(L-1) \min_{\tau \in \Gamma}\sum_{i,j\in V} \tau_{i,j} \cOT_{\cTD_{w}}{(\cT_i(T_i^L(G)), \bar{\cT}_j(T_j^L(G[S])))}.
\end{align*}

For the first term, it is easy to see that $\nu = \bm{I}$ is an optimal solution, by the same argument as in the base case: any transportation plan $\nu$ must incur the cost of transporting the excess mass $\{\norm{f^v}\}_{v \notin S}$ in the node features of $G$ which is not present in the node features of $S.$

For the second term, we can notice that $\cOT_{\cTD_{w}}{(\cT_i(T_i^L(G)), \bar{\cT}_j(T_j^L(G[S])))}$ is also an TMD between graphs which contain the corresponding multisets of depth $L-1$ trees! So, by the inductive hypothesis, $\tau = \bm{I}$ is an optimal solution for this OT problem as well. 

Consequently, substituting $\tau = \nu = \bI$, we can further simplify the lower bound to 
\begin{align*}
    \sum_{i,j \in V} \gamma_{i,j} C_{i,j} &\geq \sum_{i,j\in V} \bm{I}_{i,j} \norm{f^i - f'^j} + w(L-1) \sum_{i,j\in V} \bm{I}_{i,j} \cOT_{\cTD_{w}}{(\cT_i(T_i^L(G)), \bar{\cT}_j(T_j^L({G}[S])))} \\
    &= \sum_{i \in V}  \norm{f^i - f'^i} + w(L-1) \sum_{i \in V} \cOT_{\cTD_{w}}{\paren{\cT_i(T_i^L(G), \bar{\cT}_i(T_i^L(G[S]))}} \\
    &= \sum_{v \notin S}  \norm{f^v} + w(L-1) \sum_{v \notin S} \cOT_{\cTD_{w}}\paren{\rho{\paren{\cT_v(T_v^L(G)), \emptyset}}} \\
    &+ w(L-1) \sum_{v \in S} \cOT_{\cTD_{w}}{\paren{\cT_v(T_v^L(G)), \bar{\cT}_v(T_v^L(G[S]))}} \\
    &= \sum_{v \notin S}  \norm{f^v} + w(L-1) \sum_{v \notin S} \cOT_{\cTD_{w}}\paren{\rho\paren{\cT_v(T_v^L(G)), \emptyset}} \\
    &+ w(L-1) \sum_{v \in S} \cOT_{\cTD_{w}}\paren{\rho\paren{\cT_v(T_v^L(G)), \cT_v(T_v^L(G[S]))}}. 
\end{align*}
Finally, the inequality is tight, because $\gamma = \bI \in \Gamma$ is clearly a feasible transportation plan. So, by induction, the lemma holds. 
\end{proof}

\begin{remark}\label{remark:tmd-ot} The terms $\cOT_{\cTD_{w}}\paren{\rho\paren{\cT_v(T_v^L(G)), \cT_v(T_v^L(G[S]))}}$ in the expression of Lemma~\ref{lemma:general-decomposition} can themselves be viewed as the TMD between two graphs $\bar{G}$ and $\bar{G}'$, where $\bar{G}$ is the graph consisting of all elements in $\cT_v(T_v^L(G))$ as disjoint trees, and  $\bar{G}'$ is the graph consisting of all elements in $\cT_v(T_v^L(G[S]))$ as disjoint trees. In this sense, Lemma~\ref{lemma:general-decomposition} precisely characterizes the recursive structure of TMD between a graph and its subgraph. 
\end{remark}

\finegraineddecomp*
\begin{proof} Let $Z = \{z \in V: z \notin S\}.$ Since $T \subset S$, $T$ and $Z$ are disjoint. Moreover, $\{v \in V : v \notin S\setminus{T}\} = Z \cup T$. Thus, $Z \sqcup T = \{v \in V : v \notin S \setminus{T}\}$, where we use $\sqcup$ to denote disjoint union. So, Lemma~\ref{lemma:general-decomposition} and using the definition of $\cOTbar$ \eqref{eq:OT-bar}) shows that 
\begin{align*}
    \cTMD_{w}^L(G, G[S \setminus T]) &= \sum_{z \in Z}\norm{f^z} + w(L-1) \sum_{z \in Z} \cOT_{\cTD_{w}}\paren{\rho\paren{\cT_z(T_z^L(G)), \emptyset}} \\ 
    &+ \sum_{t \in T} \norm{f^t} + w(L-1) \sum_{t \in T} \cOT_{\cTD_{w}}\paren{\rho\paren{\cT_t(T_t^L(G)), \emptyset}}\\
    &+ w(L-1) \sum_{u \in S \setminus{T}} \cOT_{\cTD_{w}}\paren{\rho\paren{\cT_u(T_u^L(G)), \cT_u(T_u^L(G[S \setminus T]))}}.
\end{align*}
Now, each term in $\cOT_{\cTD_{w}}\paren{\rho\paren{\cT_u(T_u^L(G)), \cT_u(T_u^L(G[S \setminus T]))}}$ inside the summation is, in fact a depth $L-1$ TMD between two graphs corresponding to the two disjoint forests of computation trees (as we discussed in Remark~\ref{remark:tmd-ot}). Thus, by the inductive hypothesis, we can expand the last term as follows:
\begin{align*}
    \sum_{u \in S \setminus{T}} \cOT_{\cTD_{w}}\paren{\rho\paren{\cT_u(T_u^L(G)), \cT_u(T_u^L(G[S \setminus T]))}} &= \sum_{u \in S \setminus{T}} \cOT_{\cTD_{w}}\paren{\rho\paren{\cT_u(T_u^L(G)), \cT_u(T_u^L(G[S]))}} \\
    &+\cOT_{\cTD_{w}}\paren{\rho\paren{\cT_u(T_u^L(G[S])), \cT_u(T_u^L(G[S \setminus T]))}}.
\end{align*}
Consequently, we obtain: 
\begin{align*}
    \cTMD_{w}^L(G, G[S \setminus T]) &= \sum_{z \in Z}\norm{f^z} + w(L-1) \sum_{z \in Z} \cOT_{\cTD_{w}}\paren{\rho\paren{\cT_z(T_z^L(G)), \emptyset}} \\ 
    &+ \sum_{t \in T} \norm{f^t} + w(L-1) \sum_{t \in T} \cOT_{\cTD_{w}}\paren{\rho\paren{\cT_t(T_t^L(G)), \emptyset}}\\
    &+ w(L-1) \sum_{u \in S \setminus{T}} \cOT_{\cTD_{w}}\paren{\rho\paren{\cT_u(T_u^L(G)), \cT_u(T_u^L(G[S]))}} \\
    &+ w(L-1) \sum_{u \in S \setminus{T}}  \cOT_{\cTD_{w}}\paren{\rho\paren{\cT_u(T_u^L(G[S])), \cT_u(T_u^L(G[S \setminus T]))}}.
\end{align*}
Next, we can decompose the third summand above as follows. 
\begin{align}\label{eq:new_eq}
    \cTMD_{w}^L(G, G[S \setminus T]) &= \sum_{z \in Z}\norm{f^z} + w(L-1) \sum_{z \in Z} \cOT_{\cTD_{w}}\paren{\rho\paren{\cT_z(T_z^L(G)), \emptyset}} \\ 
    &+ \sum_{t \in T} \norm{f^t} + w(L-1) \sum_{t \in T} \cOT_{\cTD_{w}}\paren{\rho\paren{\cT_t(T_t^L(G)), \emptyset}} \notag\\
    &+ w(L-1) \sum_{u \in S} \cOT_{\cTD_{w}}\paren{\rho\paren{\cT_u(T_u^L(G)), \cT_u(T_u^L(G[S]))}} \notag \\
    &- w(L-1) \sum_{u \in T} \cOT_{\cTD_{w}}\paren{\rho\paren{\cT_u(T_u^L(G)), \cT_u(T_u^L(G[S]))}} \notag \\
    &+ w(L-1) \sum_{u \in S \setminus{T}}  \cOT_{\cTD_{w}}\paren{\rho\paren{\cT_u(T_u^L(G[S])), \cT_u(T_u^L(G[S \setminus T]))}}.\notag
\end{align}
Now, $\cOT_{\cTD_{w}}\paren{\rho\paren{\cT_u(T_u^L(G)), \cT_u(T_u^L(G[V \setminus S]))}}$ is also a depth $L-1$ TMD between two graphs corresponding to two disjoint forests of computation trees (again, see Remark~\ref{remark:tmd-ot}). So, we can apply the inductive hypothesis to see that for each $t \in T$,
\begin{align*}
    \cOT_{\cTD_{w}}\paren{\rho\paren{\cT_t(T_t^L(G)), \emptyset}} &= \cOT_{\cTD_{w}}\paren{\rho\paren{\cT_t(T_t^L(G)), \cT_t(T_t^L(G[S]))}} \\
    &+ \cOT_{\cTD_{w}}\paren{\rho\paren{\cT_t(T_t^L(G[S])), \emptyset}}. 
\end{align*}
Thus, summing over $t \in T$, we have
\begin{align*}
        \sum_{t \in T} \cOT_{\cTD_{w}}\paren{\rho\paren{\cT_t(T_t^L(G[S])), \emptyset}}   = \sum_{t \in T} \cOT_{\cTD_{w}}\paren{\rho\paren{\cT_t(T_t^L(G)), \emptyset}} - \sum_{t \in T} \cOT_{\cTD_{w}}\paren{\rho\paren{\cT_t(T_t^L(G)), \cT_t(T_t^L(G[S]))}}. 
\end{align*}
That is, by a simple rearrangement,
\begin{align*}
        \sum_{t \in T} \cOT_{\cTD_{w}}\paren{\rho\paren{\cT_t(T_t^L(G[S])), \emptyset}}   + \sum_{t \in T} \cOT_{\cTD_{w}}\paren{\rho\paren{\cT_t(T_t^L(G)), \cT_t(T_t^L(G[S]))}} = \sum_{t \in T} \cOT_{\cTD_{w}}\paren{\rho\paren{\cT_t(T_t^L(G)), \emptyset}}.
\end{align*}

By substituting this into the equation \eqref{eq:new_eq} from above, we can cancel terms to obtain 
\begin{align*}
    \cTMD_{w}^L(G, G[S \setminus T]) &= \sum_{z \in Z}\norm{f^z} + w(L-1) \sum_{z \in Z} \cOT_{\cTD_{w}}\paren{\rho\paren{\cT_z(T_z^L(G)), \emptyset}} \\ 
    &+ \sum_{t \in T} \norm{f^t} + w(L-1) \sum_{t \in T} \cOT_{\cTD_{w}}\paren{\rho\paren{\cT_t(T_t^L(G[S])), \emptyset}}\\
    &+ w(L-1) \sum_{u \in S} \cOT_{\cTD_{w}}\paren{\rho\paren{\cT_u(T_u^L(G)), \cT_u(T_u^L(G[S]))}} \\
    &+ w(L-1) \sum_{u \in S \setminus{T}}  \cOT_{\cTD_{w}}\paren{\rho\paren{\cT_u(T_u^L(G[S])), \cT_u(T_u^L(G[S \setminus T]))}}.
\end{align*}
Rearranging, we obtain 
\begin{align*}
    &\cTMD_{w}^L(G, G[S \setminus T]) = \\
    &\sum_{u \notin S}\norm{f^u} + w(L-1)\Brac{ \sum_{u \notin S} \cOT_{\cTD_{w}}\paren{\rho\paren{\cT_u(T_u^L(G)), \emptyset}} + \sum_{u \in S} \cOT_{\cTD_{w}}\paren{\rho\paren{\cT_u(T_u^L(G)), \cT_u(T_u^L(G[S]))}}} \\
    &+ \sum_{t \in T} \norm{f^t} + w(L-1) \Brac{\sum_{t \in T} \cOT_{\cTD_{w}}\paren{\rho\paren{\cT_t(T_t^L(G[S])), \emptyset}} + \sum_{t \in S \setminus{T}} \cOT_{\cTD_{w}}\paren{\rho\paren{\cT_t(T_t^L(G[S])), \cT_t(T_t^L(G[S \setminus T]))}}} \\
    &= \cTMD_{w}^L(G, G[S]) + \cTMD_{w}^L(G[S], G[S \setminus T]), 
\end{align*}
where the last equality is due to Lemma~\ref{lemma:general-decomposition} and the definition of $\cOTbar$ \eqref{eq:OT-bar}. 
\end{proof}

\algorithm*
\begin{proof} Consider Algorithm~\ref{alg:1}, TreeNorm. The runtime is immediate from the algorithm pseudocode, as matrix-vector products can be computed in $O(|E|)$. In this proof, let $A_G$ denote the adjacency matrix of a graph $G$ and let $x_G$ be the vector in $\R^{V}$ such that $x_G(v) = \norm{f^v}.$

To verify the correctness, we proceed by induction. We just need to show that for all $L$, $\cTreeNorm{G} = \norm{x_G}_1 + \norm{\sum_{\ell=1}^{L-1} \paren{\prod_{t=1}^{\ell} w(L-t)} A_G^{\ell} {x}_{G}}_1, $ as this is the exact computation computed by the algorithm pseudocode. Note that an equivalent expression is
$\cTreeNorm{G} = \norm{\sum_{\ell=0}^{L-1} \paren{\prod_{t=1}^{\ell} w(L-t)} A_G^{\ell} {x}_{G}}_1$, where the product over the emptyset is defined as 1.

If $L = 1$, then $\cTreeNorm{G}$ is simply the sum of its feature values, so this is trivially true. 

Consider the case of a depth-$L$ tree-norm computation. Throughout this proof, all variables are positive, and consequently we can move the $\ell_1$ norm in and out of sums freely (i.e., we have a triangle \emph{equality}.)

By taking $S = \emptyset$ in Lemma~\ref{lemma:general-decomposition}, we see that $\cTreeNorm{G}$ is equal to the norm of its nodes plus $w(L-1)\sum_{v \in V}\norm{{\cT_v(T_v^L(G))}}_{\text{TN}_{w}^{L-1}}.$ That is, 
\begin{align*}
    \cTreeNorm{G} = \norm{x_G}_1 + w(L-1) \sum_{v \in V} \norm{{\cT_v(T_v^L(G))}}_{\text{TN}_{w}^{L-1}}. 
\end{align*}
To interpret this expression, let $G'_v$ be the graph which is a forest of all the trees in ${\cT_v(T_v^L(G))}$-- for each $u \in N_G(v)$, we $G'_v$ will contain a depth-$(L-1)$ tree $G'_{v, u}$. Then, by inductive hypothesis,
\begin{align*}
    \norm{{\cT_v(T_v^L(G))}}_{\text{TN}_{w}^{L-1}} &= \sum_{u \in N_G(v)} \norm{{(T_v^L(G))}}_{\text{TN}_{w}^{L-1}} = \norm{G'_v}_{\text{TN}_w^{L-1}} = \sum_{u \in N_G(v)} \norm{G'_{u,v}}_{\text{TN}_w^{L-1}} \\
    &= \sum_{u \in N_G(v)} \norm{\sum_{\ell=0}^{L-2} \paren{\prod_{t=1}^\ell w(L-t-1)} A_{G_{u,v}}^{\ell} x_{G_{u,v}}}_1 \\
    &=  \sum_{u \in N_G(v)} \sum_{\ell=0}^{L-2} \paren{\prod_{t=1}^\ell w(L-t-1)} \norm{A_{G_{u,v}}^{\ell} x_{G_{u,v}}}_1,  
\end{align*}
Letting $A_G(u, v)$ be the $(u, v)$-th entry of $A_G$, we can turn the sum over $N_G$ into a sum over $V$ by taking an additional product with the adjacency matrix $A_G$: 
\begin{align*}
    \norm{{\cT_v(T_v^L(G))}}_{\text{TN}_{w}^{L-1}} &= \sum_{u \in N_G(v)} \sum_{\ell=0}^{L-2} \paren{\prod_{t=1}^\ell w(L-t-1)} \norm{A_{G_{u,v}}^{\ell} x_{G_{u,v}}}_1 \\
    &= \norm{ \sum_{u \in V} \sum_{\ell=0}^{L-2} \paren{\prod_{t=1}^\ell w(L-t-1)} A_G(u, v) A_{G_{u,v}}^{\ell} x_{G_{u,v}}}_1. 
\end{align*}
Now, by the way we constructed the trees ${\cT_v(T_v^L(G))}$, observe that $u$ has the exact same $(L-1)$-hop neighborhood in $G$ as it does in $G_{u,v}$. Consequently,
\begin{align*}
    \cTreeNorm{G} &= \norm{x_G}_1 + w(L-1) \sum_{v \in V} \norm{{\cT_v(T_v^L(G))}}_{\text{TN}_{w}^{L-1}} \\
    &= \norm{x_G}_1 + w(L-1) \norm{  \sum_{\ell=0}^{L-2} \paren{\prod_{t=1}^\ell w(L-t-1)} \sum_{v \in V} \sum_{u \in V} A_G(u, v) A_{G_{u,v}}^{\ell} x_{G_{u,v}}}_1 \\
    &= \norm{x_G}_1 + \norm{\sum_{\ell=1}^{L-2}  \paren{\prod_{t=1}^\ell w(L-t-1)} \sum_{v \in V}A_G^{\ell+1} {x}_{G}}_1 \\
    &= \norm{x_G}_1 + \norm{\sum_{\ell=1}^{L-1} \paren{\prod_{t=1}^{\ell} w(L-t)} A_G^{\ell} {x}_{G}}_1, 
\end{align*}
completing the induction and the proof of correctness. 
\end{proof}

\subsection{Proof of NP-completeness of node subsampling problem \eqref{eq:node-subsampling}} \label{sec:hardness}

Finally, we prove that optimizing \eqref{eq:node-subsampling} is NP-complete. Concretely, by Theorem~\ref{thm:node-subsampling-relaxed}, we have \eqref{eq:node-subsampling} is equivalent to \eqref{eq:final-opt-problem} for $\cS = 2^V$. Consequently, it suffices to show that \eqref{eq:final-opt-problem} is NP-complete for $\cS = 2^V$.

As is standard in NP-completeness proofs, we consider the following decision version of the problem, for any \emph{arbitrary} positive weights $w$. 

\begin{definition}[Tree norm decision problem]\label{def:decision-problem} In the tree-norm decision problem, we are given $(G, L, k, \tau)$ where $G = (V, E, f)$ is a node-weighted graph, $k, L \geq 1$ are integers, and $\tau > 0$ is a threshold. We must output YES if $\max_{S \in 2^V; \abs{S} = k} \cTreeNorm{G[S]} \geq \tau$, and output NO otherwise. 
\end{definition}

The proof will proceed by reduction to $k$-clique, which is a well-established NP-hard problem. 
\begin{definition}[$k$-clique problem]\label{def:k-clique} In the $k$-clique problem, we are given an \emph{unweighted} graph $G = (V, E)$ and integer $k \geq 1$. We must output YES if $G$ contains a $k$-clique and NO otherwise. 
\end{definition}

\begin{theorem}\label{thm:np-hard} Solving the tree norm decision problem (Problem~\ref{def:decision-problem}) is NP-complete. 
\end{theorem} 

\begin{proof} To show that the problem is in NP, note that given a candidate subset $S \subset V$ as a witness, we can compute the tree norm $\cTreeNorm{G[S]}$ in polynomial time (e.g., using Algorithm~\ref{alg:1}), and verify whether or not $\cTreeNorm{G[S]} \geq \tau$ in additional constant time. Thus, the problem is polynomial-time verifiable, and thus in NP. 

Next, we will prove that the problem is NP hard, by reduction to $k$-clique. Suppose, there exists a polynomial time algorithm for solving the tree norm decision problem. Then, will show that there exists a polynomial time algorithm for the $k$-clique problem. 

So, suppose we are given an unweighted graph $G = (V, E)$ and $k$ as input. Let $\bar{G} = (V, E, g)$ where $g^v = 1 \in \R^d$ for all $v \in V$. Let $K_k = (V_k, E_k)$ denote a $k$-clique graph, and let $\bar{K}_K \defeq (V_k, E_k, f)$ where $f^v = 1 \in \R$ for all $v \in V_k$. That is, $\bar{K}_K$ is just a $k$-clique in which every node is assigned a 1-dimensional feature value of 1, and $\bar{G}$ is just the original input graph $G$ in which every node is assigned a 1-dimensional feature value of 1.

Consider the algorithm which does the following: 
\begin{enumerate}
    \item Compute $\tau_{k\mathrm{-clique}} \defeq \norm{\bar{K}_k}_{\mathrm{TN}_w^k}$. This can be computed in polynomial time (e.g., using Algorithm~\ref{alg:1}.)
    \item Solve and output the solution to the tree norm decision for $(\bar{G}, k, k, \tau_{k\mathrm{-clique}})$. 
\end{enumerate}

We will show that this procedure solves the $k$-clique problem. Suppose that the algorithm outputs NO. Then, clearly, $G$ does not contain a $k$-clique, or else, $\bar{G}$ would contain $\bar{K}_k$ as a subgraph and thus, 
\begin{align*}
    \max_{S \in 2^V; \abs{S} = k} \norm{G[S]}_{\mathrm{TN}_w^k} \geq \norm{\bar{K}_k}_{\mathrm{TN}_w^k} = \tau_{k\mathrm{-clique}}, 
\end{align*}
which is a contradiction. 

Now, suppose the algorithm outputs YES. Then, $\bar{G}$ contains a $k$-subgraph with tree norm greater than or equal to $\tau_{k\mathrm{-clique}}$. However, from the definition of the tree norm, it is easy to see that on a graph where all nodes have feature value 1 and the weight function $w > 0$ is positive, the tree-norm $\norm{G[S]}_{\mathrm{TN}_w^k}$ is \emph{monotonic} as a function of the number of edges in the multiset of depth-$k$ computation trees defined by $G[S]$ (see Definition~\ref{def:tree}.) Thus, 
\begin{align*}
    \max_{S \in 2^V; \abs{S} = k} \norm{G[S]}_{\mathrm{TN}_w^k} \leq \bar{K}_k, 
\end{align*}
since $\bar{K}_k$ is the \emph{unique} subgraph on $k$-nodes which contains \emph{every possible} edge in its depth-$k$ computation trees. And consequently, the algorithm outputs YES if and only if $\bar{G}$ contains $\bar{K}_k$ as a $k$-clique.  
\end{proof}

\end{document}